\documentclass{ieeeaccess}
\usepackage{cite}
\usepackage{amsmath,amssymb,amsfonts}
\usepackage{algorithmic}
\usepackage{graphicx}
\usepackage{textcomp}
\usepackage[utf8]{inputenc}
\usepackage[T1]{fontenc} 
\usepackage{newtxmath}
\usepackage{newtxtext}   
\usepackage{helvet}      
\usepackage{courier}     
\usepackage{multirow}
\usepackage{geometry}
\usepackage{booktabs}
\usepackage{array}
\usepackage{mathtools}
\usepackage{bm}
\usepackage{physics}
\usepackage{caption}
\usepackage{subcaption}
\usepackage{enumitem}
\usepackage{graphicx}
\usepackage{float}

\newcommand{\R}{\mathbb{R}}
\newcommand{\C}{\mathbb{C}}

\def\BibTeX{{\rm B\kern-.05em{\sc i\kern-.025em b}\kern-.08em
    T\kern-.1667em\lower.7ex\hbox{E}\kern-.125emX}}
\usepackage[colorlinks=true, linkcolor=blue, citecolor=blue, urlcolor=blue]{hyperref}
\makeatletter

\newlength{\xfigwd}
\makeatother
\UseRawInputEncoding

\begin{document}
\history{Date of publication xxxx 00, 0000, date of current version xxxx 00, 0000.}
\doi{10.1109/TQE.2020.DOI}

\title{Quantum Temporal Convolutional Neural Networks for Cross-Sectional Equity Return Prediction: A Comparative Benchmark Study}
\author{
\uppercase{Chi-Sheng Chen}\authorrefmark{1}\authorrefmark{*},
\uppercase{Xinyu Zhang}\authorrefmark{2}\authorrefmark{*},
\uppercase{En-Jui Kuo}\authorrefmark{5},
\uppercase{Rong Fu}\authorrefmark{2},
\uppercase{Qiuzhe Xie}\authorrefmark{3},
and \uppercase{Fan Zhang}\authorrefmark{4}
}

\address[1]{Beth Israel Deaconess Medical Center \& Harvard Medical School, Boston, MA 02115 USA}
\address[2]{Luddy School of Informatics, Computing, and Engineering, Indiana University Bloomington, Bloomington, IN 47405 USA}
\address[3]{Institute of Electronics Engineering, National Taiwan University, 106319, Taiwan}
\address[4]{Department of Mathematics, Boise State University, Boise, ID 83702 USA}
\address[5]{Department of Electrophysics, National Yang Ming Chiao Tung University, 30010,Taiwan}
\address[*]{Chi-Sheng Chen and Xinyu Zhang contributed equally to this work}

\markboth
{Chen \headeretal: Preparation of Papers for IEEE Transactions on Quantum Engineering}
{Chen \headeretal: Preparation of Papers for IEEE Transactions on Quantum Engineering}

\corresp{Corresponding author: Fan Zhang (email: fanzhang@boisestate.edu)}

\begin{abstract}
Quantum machine learning offers a promising pathway for enhancing stock market prediction, particularly under complex, noisy, and highly dynamic financial environments. However, many classical forecasting models struggle with noisy input, regime shifts, and limited generalization capacity. To address these challenges, we propose a Quantum Temporal Convolutional Neural Network (QTCNN) that combines a classical temporal encoder with parameter-efficient quantum convolution circuits for cross-sectional equity return prediction. The temporal encoder extracts multi-scale patterns from sequential technical indicators, while the quantum processing leverages superposition and entanglement to enhance feature representation and suppress overfitting. We conduct a comprehensive benchmarking study on the JPX Tokyo Stock Exchange dataset and evaluate predictions through long-short portfolio construction using out-of-sample Sharpe ratio as the primary performance metric. QTCNN achieves a Sharpe ratio of 0.538, outperforming the best classical baseline by approximately 72\%. These results highlight the practical potential of quantum-enhanced forecasting model, QTCNN, for robust decision-making in quantitative finance.
\end{abstract}


\titlepgskip=-15pt

\maketitle

\section{Introduction}
\label{sec:introduction}

Stock market forecasting is a major challenge in computational finance\cite{10545224}. Financial markets produce huge amounts of complex data every day. This data is noisy, changes constantly, and contains many hidden patterns\cite{10862401}. Creating models that accurately predict market movements is highly valuable, as strong forecasts enable better investment strategies and more effective risk management. 

In recent years, researchers have used many different computational methods to predict stock prices. These include classic statistical models \cite{11115662}, machine learning algorithms, and modern deep learning methods like Transformers \cite{9357288}. While these methods have shown some success, they often face significant problems. Many models struggle to capture the complex, long-term dynamics of the market. They may also overfit to past data, which means they perform poorly when market conditions change\cite{11140787}. As a result, many existing methods lack the robustness and adaptability needed for real-world trading.
\begin{table*}[t]
\centering
\caption{Comparison of Recent Stock Prediction Methods}
\label{TABLE 1}
\footnotesize
\setlength{\tabcolsep}{3pt}
\renewcommand{\arraystretch}{1.15}
\begin{tabular}{p{2.5cm} p{3.2cm} p{2.6cm} p{3.2cm} p{2.5cm}}
\toprule
\textbf{Study} & \textbf{Method / Model} & \textbf{Target Data} & \textbf{Key Techniques} & \textbf{Performance} \\
\midrule

A{\c{s}}{\i}r{\i}m, et al.\cite{11171624} 
& Single-layer Transformer 
& Nasdaq daily closing prices 
& Data segmentation, independent normalization, sinusoidal positional embeddings 
& Average correlation $\approx 0.96$, effectively captured stock price trends \\

Ahire et al.\cite{10882429} 
& Hybrid ARIMA + Machine Learning 
& ICICI and SBI stock data 
& ARIMA with RSI, SMA, and Stochastic K indicators 
& Prediction accuracy $\approx 0.96$ (SBI) and $0.94$ (ICICI); effective bullish/bearish detection \\

Singh et al.\cite{10545132} 
& Random Forest, Gradient Boosting 
& TCS, HCL, and Ques Corp (5-year data) 
& Multi-model performance comparison 
& $R^2 > 0.99$ and MAPE $< 0.2\%$, highest accuracy \\

RaviTeja et al.\cite{10690751} 
& Novel Quantum Enforcing Algorithm (NQEA) 
& Stock index and Bitcoin prices 
& Quantum-inspired optimization, compared with BV algorithm 
& Accuracy $95.9\%$, loss $4.1\%$, outperforming BV (88.0\%) \\

Hossain et al.\cite{11171862} 
& Linear Regression, RF, GB, XGBoost, MLP 
& Dhaka Stock Exchange 
& Cross-model performance comparison 
& Linear Regression achieved lowest prediction error and most consistent performance \\

\bottomrule
\end{tabular}
\end{table*}
Quantum computing offers a new and promising approach to these challenges\cite{10461108}. Quantum Machine Learning (QML) uses the principles of quantum mechanics, such as superposition and entanglement. These features allow quantum models to explore very large, complex data spaces\cite{10864082}. This may help them find important patterns in financial data that classical computers cannot.

Recent studies (shown as TABLE~\ref{TABLE 1}) have explored a diverse range of computational and intelligent approaches to improve the accuracy and efficiency of stock market forecasting. These works span from deep learning architectures and hybrid statistical-machine learning frameworks to quantum-inspired algorithms.

Despite their notable results, the studies above share several common weaknesses. Most relied on limited dataset quality, which restricted generalization across diverse market conditions. The Transformer and several traditional machine learning approaches still struggled to capture nonlinear and long-term dependencies within financial time-series data, often showing risks of overfitting and limited incorporation of richer signals such as macroeconomic or sentiment-driven factors.

In this paper, we propose a quantum-enhanced approach to address these limitations. Our model QTCNN, which leverages superposition and entanglement to extract high-dimensional, nonlinear relationships within noisy market data. This architecture is designed to improve generalization and boost robustness compared with classical models by providing richer representational capacity while maintaining computational efficiency. 

To validate our model, we conduct a comprehensive benchmarking study using the JPX Tokyo Stock Exchange dataset\cite{KaggleJPX2025}. We compare our proposed QTCNN model against a strong set of classical baselines, including LightGBM (LGBM)\cite{NIPS2017_6449f44a}, LSTM\cite{11064676}, and the Transformer\cite{10920805}. We also compare it to other recently published quantum-inspired algorithms. Our results show that our model achieves superior predictive accuracy and robustness, demonstrating a more effective way to handle complex financial time-series data.

The main contributions of this paper can be concluded as bellow:

\begin{itemize}
\item 
We proposed an end-to-end quantum-enhanced framework for cross-sectional equity return prediction, where the QNN serves as the primary predictive model to extract complex market signals.

\item 
The QNN leverages quantum superposition and entanglement to model nonlinear, high-dimensional structures in noisy financial data, improving generalization capability while mitigating overfitting compared with classical machine learning approaches.

\item
A unified benchmarking study is conducted against representative deep learning architectures, statistical methods, and other quantum-based models, demonstrating the superior predictive accuracy and robustness of the proposed framework in a realistic financial forecasting scenario.
\end{itemize}

The rest of this article is organized as follows. \textbf{Section II} reviews the task and dataset used in this work. \textbf{Section III} details the architecture and provides a quantum resource complexity analysis of our proposed QTCNN model. \textbf{Section IV} presents numerical results including a qubit scalability study, mechanism analysis of quantum enhancement, and noise resilience experiments. Finally, \textbf{Section V} discusses dequantization considerations, prospects for NISQ and fault-tolerant execution, and future research directions.

\begin{table}[t]
\centering
\caption{Fields in the Price Data Table 
\label{TABLE 2}
(\texttt{stock\_prices})}
\footnotesize
\setlength{\tabcolsep}{4pt}
\renewcommand{\arraystretch}{1.15}
\begin{tabular}{p{2.8cm} p{3.8cm}}
\toprule
\textbf{Field Name} & \textbf{Description} \\
\midrule

Date & Trading date. \\

SecuritiesCode & Unique stock identifier (e.g., 1301, 1332), used as a key together with Date. \\

Open & Daily opening price. \\

High & Daily highest price. \\

Low & Daily lowest price. \\

Close & Daily closing price. \\

Volume & Daily traded volume (in shares). \\

AdjustmentFactor & Cumulative adjustment factor accounting for stock splits, reverse splits, dividends, and other corporate actions. \\

SupervisionFlag & Indicator for stocks under exchange supervision or at risk of delisting. \\

Target & Official competition target; two-day forward return based on future prices (see Section II-E). \\

\bottomrule
\end{tabular}
\end{table}

\begin{table}[t]
\centering
\caption{Fields in the Static Stock Information Table (\texttt{stock\_list})}
\label{TABLE 3}
\footnotesize
\setlength{\tabcolsep}{4pt}
\renewcommand{\arraystretch}{1.15}
\begin{tabular}{p{2.8cm} p{3.8cm}}
\toprule
\textbf{Field Name} & \textbf{Description} \\
\midrule

SecuritiesCode & Stock identifier matching \texttt{stock\_prices}. \\

Name & Official stock name. \\

MarketCapitalization & Market capitalization of the company. \\

Universe0 & Boolean flag indicating whether the stock belongs to the core prediction universe. \\

\bottomrule
\end{tabular}
\end{table}

\begin{figure}[t]
    \centering
    \includegraphics[width=\linewidth]{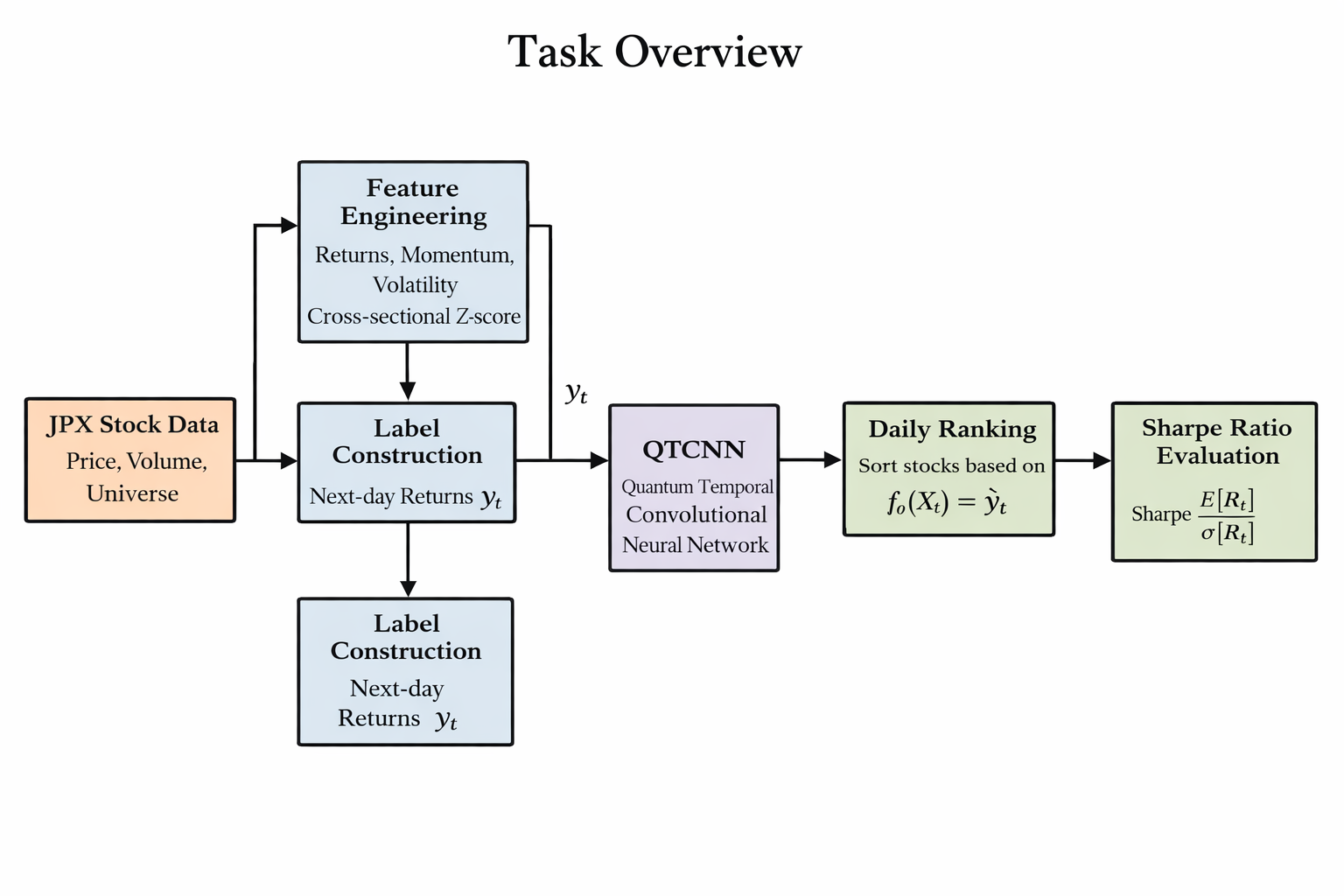}
    \caption{Overview of the Stock Prediction Task and Portfolio Evaluation Pipeline.}
    \label{Task Overview}
\end{figure}

\section{Task and Dataset}

This section describes the official dataset of the Kaggle "\href{https://www.kaggle.com/competitions/jpx-tokyo-stock-exchange-prediction}{JPX Tokyo Stock Exchange Prediction}" competition and how it is used in our study, including the construction of training labels and subset sampling schemes for computationally efficient yet reproducible experiments.

\subsection{Task Overview}

FIGURE~\ref{Task Overview} illustrates the overall pipeline of the stock prediction task and portfolio evaluation framework used in this study. The process consists of data preparation, feature and label construction, model prediction, cross-sectional ranking, and portfolio performance evaluation.

\textbf{Data and features.}
We begin with the JPX stock dataset, which provides daily price and volume information for all listed securities within the official prediction universe. From the raw price series, we construct a set of technical and statistical features, including returns, momentum indicators, volatility measures, and price--volume related variables. To ensure comparability across stocks on the same trading day, all features are standardized in a cross-sectional manner (within each day) using z-score normalization.

\textbf{Label construction.}
For each stock $k$ on trading day $t$, the supervision signal is derived from the official competition target, which corresponds to the forward return over the next holding period. These targets serve as the ground-truth signals that guide the model to learn cross-sectional differences in expected future performance.

\textbf{Model prediction.}
The engineered features for each day are fed into the predictive model (e.g., QTCNN or other baseline models), which outputs a continuous score $\hat{y}_t^{(k)}$ for every stock. This score reflects the model's assessment of the relative attractiveness of each stock for the upcoming holding period.

\textbf{Cross-sectional ranking and portfolio formation.}
On each trading day, stocks are sorted based on the predicted scores $\hat{y}_t^{(k)}$. A long--short portfolio is then constructed by going long the top $N$ ranked stocks and short the bottom $N$ ranked stocks, resulting in a market-neutral portfolio that captures the cross-sectional signal extracted by the model.

\textbf{Performance evaluation.}
The daily returns of the long--short portfolio are aggregated over the out-of-sample period to assess predictive performance. The primary evaluation metric is the Sharpe ratio, defined as the mean portfolio return divided by its standard deviation, which measures the risk-adjusted effectiveness of the proposed prediction and ranking pipeline.

\subsection{Original Dataset and Stock Universe}

\subsubsection{Data source and time span}

The dataset is provided by JPX via the Kaggle competition. The official training data (``train'' stock prices) covers approximately 2017--2021 Japanese trading days, and the supplemental files extend the timeline into 2022. On each trading day, there are roughly $\sim 2000$ ``active'' stocks with complete price, volume, and target information.

\subsubsection{Universe definition and \texttt{Universe0}}

In addition to price data, JPX provides a static stock information table, denoted as \texttt{stock\_list}, which contains a key flag \texttt{Universe0}:

\begin{itemize}
    \item \texttt{Universe0 = 1} (or \texttt{True}) identifies stocks that belong to a core universe of high-capitalization, sufficiently liquid names (approximately the top $\sim 2000$ stocks by size and liquidity).
    \item These \texttt{Universe0} stocks constitute the official prediction universe in the competition.
\end{itemize}

In our study, we focus on stocks with \texttt{Universe0 = True} to match the official setting and to avoid extremely illiquid securities.

\subsection{Key Fields and Financial Interpretation}

\subsubsection{Price data (\texttt{stock\_prices})}

The main price dataset, available in both the training and supplemental sets, includes the fields summarized in TABLE~\ref{TABLE 2}.

These fields form the raw basis for our feature engineering and financial interpretation.

\subsubsection{Static stock information (\texttt{stock\_list})}

The \texttt{stock\_list} table contains relatively static attributes for each stock, as summarized in Table~\ref{TABLE 3}.

In the present work, we primarily use \texttt{Universe0} to define the prediction universe. Sector codes are not used as input features in the main models but can be exploited for industry-neutral analysis or robustness checks.

\subsubsection{Adjusted close price}

To eliminate non-economic jumps in price series caused by corporate actions, we construct an \emph{adjusted close} price for each stock $k$ at time $t$ using the cumulative adjustment factors:

\begin{equation}
\mathrm{AdjClose}_{k,t}
= \mathrm{Close}_{k,t}
\cdot \prod_{\tau \le t} \mathrm{AdjustmentFactor}_{k,\tau}.
\end{equation}

Here, $k$ indexes the stock, and $t$ indexes the trading day. The adjusted close $\mathrm{AdjClose}_{k,t}$ is used to compute returns, momentum, volatility, and other time-series features.

Importantly, the official \texttt{Target} remains defined on the \emph{raw} closes; we do not modify the target. Adjustments are applied only on the feature side.

\subsection{Definition and Interpretation of the Target}
\label{subsec:target_def}

The \texttt{Target} field is the central label of the JPX competition and corresponds to a \emph{two-day holding return with a one-day execution delay}.

Let $C(k,t)$ denote the raw close price of stock $k$ at trading day $t$. The official target can be written as:
\begin{equation}
\mathrm{Target}(k,t)
= \frac{C(k,t+2) - C(k,t+1)}{C(k,t+1)}.
\end{equation}

Financially, this corresponds to the following strategy:

\begin{itemize}
    \item On day $t$, the model observes information up to and including day $t$ and produces a ranking for day $t$.
    \item At the close of day $t+1$, a position is opened at price $C(k,t+1)$ (long or short depending on the ranking).
    \item At the close of day $t+2$, the position is closed at price $C(k,t+2)$.
    \item The realized return is exactly $\mathrm{Target}(k,t)$ as defined above.
\end{itemize}

Thus, the model at day $t$ predicts the return of a two-day holding strategy executed over $[t+1, t+2]$. This design enables daily rebalancing while avoiding intraday trading assumptions.

\subsection{Derived Features and Preprocessing}

\subsubsection{Single-stock time-series features}

Starting from $\mathrm{AdjClose}$, we compute for each stock $k$ a set of univariate time-series features:

\begin{itemize}
    \item \textbf{Returns and momentum}
    \begin{itemize}
        \item One-period return:
              \begin{equation}
              \mathrm{ret1}_{k,t}
              =
              \dfrac{\mathrm{AdjClose}_{k,t} - \mathrm{AdjClose}_{k,t-1}}{\mathrm{AdjClose}_{k,t-1}}.
              \end{equation}
        \item Momentum indicators $\mathrm{mom}\{1, 2, 5, 10, 20\}$, e.g., cumulative return over the past $n$ days or deviations from moving averages.
    \end{itemize}
    \item \textbf{Volatility features}
    \begin{itemize}
        \item Rolling standard deviation of returns: $\mathrm{vol}\{5, 10, 20\}$ computed from $\mathrm{ret1}$ over windows of length $5$, $10$, and $20$ days.
    \end{itemize}
    \item \textbf{Price--volume features}
    \begin{itemize}
        \item Dollar volume:
              \(
              \mathrm{dv}_{k,t} = \mathrm{Close}_{k,t} \times \mathrm{Volume}_{k,t}.
              \)
        \item Rolling means of volume and dollar volume: $\mathrm{dv\_mean}\{5, 10, 20\}$, $\mathrm{vol\_mean}\{5, 10, 20\}$.
        \item Log-transformed features such as 
        
        $\log(\mathrm{Volume}_{k,t})$ 
        and $\log(\mathrm{dv}_{k,t})$ to mitigate heavy-tailed distributions.
    \end{itemize}
\end{itemize}

\subsubsection{Cross-sectional normalization}

For each trading day $t$, we apply a cross-sectional preprocessing pipeline over all stocks (typically restricted to \texttt{Universe0}):

\begin{enumerate}
    \item \textbf{Winsorization.} For each feature, values are winsorized at the $1$st and $99$th percentiles across stocks on day $t$ to trim extreme outliers.
    \item \textbf{Z-score standardization.} For winsorized feature values $\{x_{k,t}\}_k$, we compute
    \begin{equation}
        z_{k,t} = \frac{x_{k,t} - \mu_t}{\sigma_t},
    \end{equation}
    
    where $\mu_t$ and $\sigma_t$ are the cross-sectional mean and standard deviation of that feature on day $t$.
\end{enumerate}

This procedure emphasizes \emph{relative} signals within each day's cross-section rather than absolute price or volume levels, aligning with the cross-sectional stock selection framework.

\subsection{Label Construction and Weakly Supervised Training Samples}

Although the official \texttt{Target} is a continuous value, we adopt a weakly supervised scheme that focuses on the most clearly positive and negative examples.

For each trading day $t$:

\begin{enumerate}
    \item We rank all stocks by $\mathrm{Target}(k,t)$ in descending order.
    \item We select:
    \begin{itemize}
        \item The top $p$ stocks as the \emph{positive} class with binary label $\mathrm{label\_bin}(k,t) = 1$;
        \item The bottom $p$ stocks as the \emph{negative} class with binary label $\mathrm{label\_bin}(k,t) = 0$.
    \end{itemize}
    \item Stocks in the middle range are excluded from training and treated as low-signal observations.
\end{enumerate}

In our experiments, $p$ is typically set to $200$ or $300$. This construction increases the signal-to-noise ratio by concentrating supervision on the strongest long and short candidates and mirrors the eventual long--short portfolio structure used in evaluation.

\subsection{Temporal Split and Sequence Construction}

\subsubsection{Temporal train--test split}

To avoid look-ahead bias, we perform a simple chronological split:

\begin{itemize}
    \item After sorting all days by \texttt{Date}, the first $80\%$ of days constitute the training period.
    \item The remaining $20\%$ of days constitute the out-of-sample test period.
\end{itemize}

All model training and hyperparameter tuning are conducted on the training period. The Sharpe ratios reported in our results are computed solely on the OOS period.

\subsubsection{Sequence construction for time-series models}

For sequence models such as LSTMs, Transformers, or quantum-enhanced variants, we construct input sequences of fixed length $\mathrm{seq\_len}$ (default $\mathrm{seq\_len} = 20$):

\begin{equation}
\mathbf{x}_{k,t}
=
\big[
\mathrm{features}_{k,t-19},
\ldots,
\mathrm{features}_{k,t}
\big],
\end{equation}

where $\mathrm{features}_{k,\tau}$ denotes the feature vector for stock $k$ at day $\tau$. Each sequence $\mathbf{x}_{k,t}$ is paired with the label $\mathrm{Target}(k,t)$ or its binary counterpart $\mathrm{label\_bin}(k,t)$.

\subsection{Subset Sampling of Trading Days}

To control computational cost and facilitate fair comparisons across different model architectures, we design two sampling schemes at the \emph{trading-day} level. In both schemes, sampling is applied \emph{before} feature engineering and sequence construction.

\subsubsection{Stride sampling}

Let $\{t_1, t_2, \ldots, t_T\}$ be the chronologically ordered trading days. Given a stride $k$ (e.g., $k = 11$), we select the subset

\begin{equation}
\{t_1, t_{1+k}, t_{1+2k}, \ldots \}.
\end{equation}

This ``stride'' subsampling:

\begin{itemize}
    \item Preserves the overall temporal span (e.g., 2017--2021) while sparsifying the calendar.
    \item Uses approximately $1/k$ of all trading days (for $k=11$, about $9\%$ of the original days).
    \item Substantially reduces data volume and training time, making it suitable for rapid prototyping and architectural exploration.
\end{itemize}

\subsubsection{Year-wise stratified fraction sampling}

Alternatively, we partition trading days by calendar year (e.g., 2017, 2018, \dots, 2021) and perform stratified random sampling:

\begin{enumerate}
    \item For each year, we randomly sample a fixed fraction $\alpha$ (e.g., $\alpha = 0.1$) of trading days from that year.
    \item A fixed random seed is used to ensure reproducibility.
\end{enumerate}

This year-wise stratification:

\begin{itemize}
    \item Ensures that different market regimes (bull, sideways, crisis periods, etc.) are proportionally represented in the subset.
    \item Avoids concentrating the subset on a single year with exceptionally high volatility or data density.
\end{itemize}

\subsubsection{Relation between subsets and the full dataset}

All subsequent steps---feature engineering, label construction, temporal splitting, and sequence construction---are performed on the selected subset of trading days. Using the full dataset is equivalent to setting the stride to $k=1$ or the fraction to $\alpha = 1$ (i.e., no subsampling).

These elements fully specify how the original JPX dataset is transformed into the training and evaluation sets used in our experiments.

\section{Methodology}
\subsection{Quantum State Representation}

An $n$-qubit quantum state resides in the Hilbert space $\mathcal{H} = (\C^2)^{\otimes n}$ with dimension $2^n$. A pure state is represented as:
\begin{equation}
    \ket{\psi} = \sum_{j=0}^{2^n-1} \alpha_j \ket{j}, \quad \sum_{j=0}^{2^n-1}|\alpha_j|^2 = 1
    \label{eq:quantum_state}
\end{equation}
where $\{\ket{j}\}$ forms the computational basis \cite{nielsen2010quantum}.

\subsection{Parameterized Quantum Circuits}

Variational quantum algorithms employ parameterized unitary operations $U(\bm{\theta})$ where $\bm{\theta} \in \R^p$ are trainable parameters \cite{cerezo2021variational}:
\begin{equation}
    U(\bm{\theta}) = \prod_{l=1}^{L} W_l U_l(\bm{\theta}_l)
    \label{eq:pqc}
\end{equation}
where $W_l$ are fixed entangling operations and $U_l(\bm{\theta}_l)$ are parameterized rotations.

\subsection{Feature Encoding}

Classical data $\bm{x} \in \R^n$ is encoded into quantum states via angle embedding \cite{schuld2021machine}:
\begin{equation}
    S(\bm{x}) = \bigotimes_{j=1}^{n} R_Y(x_j) \ket{0}^{\otimes n}
    \label{eq:angle_embed}
\end{equation}
where $R_Y(\theta) = \exp(-i\theta Y/2)$ is the Pauli-Y rotation gate.

\subsection{Measurement and Expectation Values}

Predictions are obtained by measuring Pauli observables:
\begin{equation}
    \langle O \rangle = \bra{\psi(\bm{x}, \bm{\theta})} O \ket{\psi(\bm{x}, \bm{\theta})}
    \label{eq:expectation}
\end{equation}
For binary classification, the Pauli-Z expectation $\langle Z \rangle \in [-1, 1]$ on designated qubits provides the model output.


\begin{figure*}[t]
    \centering
    \includegraphics[width=\textwidth]{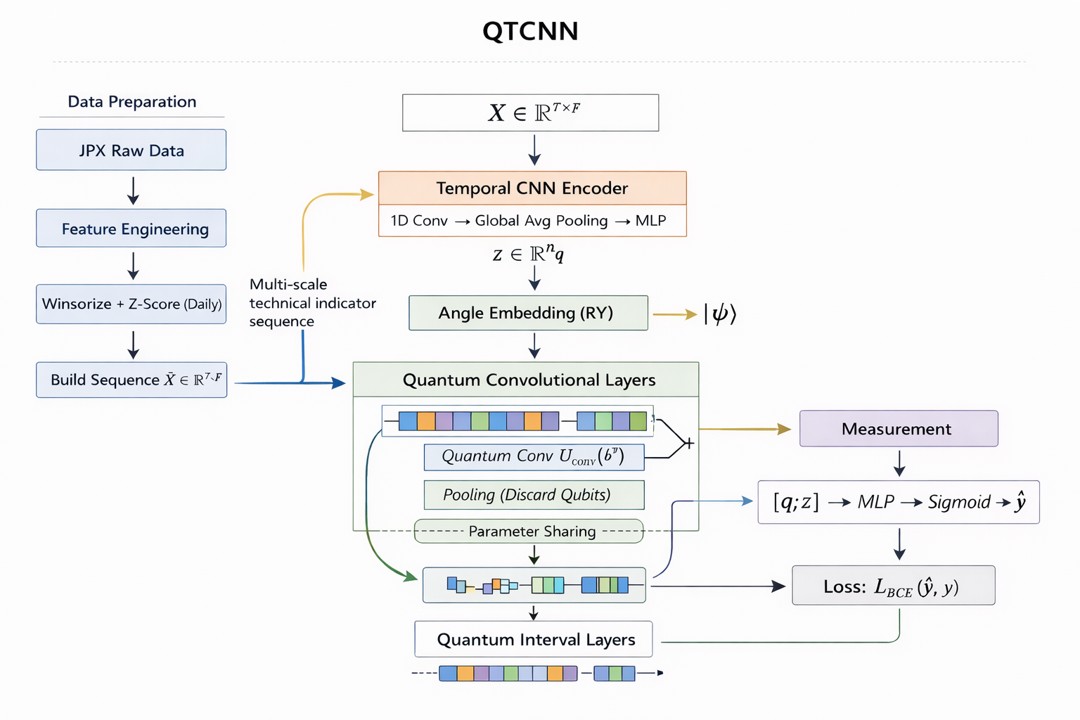}
    \caption{Architecture of the Proposed QTCNN Model.}
    \label{QTCNN}
\end{figure*}

\subsection{Quantum Temporal Convolutional Neural Network (QTCNN)}
The central methodological contribution of this work is the 
\textbf{Quantum Temporal Convolutional Neural Network (QTCNN)}(shown as FIGURE~\ref{QTCNN} ), 
which addresses two key limitations of existing quantum approaches 
for sequential data:

\begin{enumerate}
    \item \textbf{Temporal feature extraction:} Unlike standard QCNN 
    that applies PCA to flatten temporal structure, QTCNN employs a 
    classical temporal encoder to preserve sequential dynamics before 
    quantum processing.
    
    \item \textbf{Parameter efficiency:} Through intra-layer parameter 
    sharing, QTCNN reduces quantum circuit parameters from $\Theta(6 \cdot \sum_{l=1}^{L} \lfloor n_q/2^l \rfloor) = \Theta(6(n_q-1))$ in conventional QCNN (for $n_q=2^k$) to $\Theta(6 \cdot L_{\text{eff}})$ where $L_{\text{eff}}=\lfloor \log_2(n_q) \rfloor$. For instance, at $n_q=8$, this reduces the quantum parameter count from 42 to 18 (a 2.3$\times$ compression), mitigating trainability issues associated with barren plateaus in deep variational circuits.
\end{enumerate}

We introduce the QTCNN, a hybrid quantum-classical architecture specifically tailored for financial time series data. The QTCNN architecture implements quantum convolutional operations inspired by classical CNNs \cite{cong2019quantum, henderson2020quanvolutional}.

The remaining quantum architectures (QNN, QCNN, QLSTM, QASA, QSVM, 
PQC+LGBM) serve as \textit{ablation baselines} to isolate the 
contribution of each design choice.

\subsubsection{Input Processing}

Tabular features $\bm{x} \in \R^d$ undergo dimensionality reduction:
\begin{equation}
    \bm{z} = W_{\text{PCA}} \cdot \text{StandardScaler}(\bm{x}) \in \R^{n_q}
    \label{eq:qtcnn_input}
\end{equation}
where $n_q = 8$ qubits and $W_{\text{PCA}} \in \R^{n_q \times d}$ is the PCA projection matrix.

\subsubsection{Quantum Convolutional Layer}

The QTCNN circuit applies hierarchical convolution-pooling operations. The convolutional unitary on adjacent qubits $(i, i+1)$ is:
\begin{equation}
\begin{aligned}
    U_{\text{conv}}(\bm{\theta}) = \text{CNOT}_{i,i+1} \cdot (R_Y(\theta_1) \otimes R_Z(\theta_2)) \cdot \\ \text{CNOT}_{i+1,i} \cdot (R_Y(\theta_3) \otimes R_Z(\theta_4)) \cdot (R_Y(\theta_5) \otimes R_Z(\theta_6))
    \label{eq:qcnn_conv}
\end{aligned}
\end{equation}
with 6 parameters per convolutional unit. After each convolutional layer, pooling is performed by measuring and discarding alternating qubits, reducing the circuit width by half.

\subsubsection{Hybrid Classification Head}

The quantum output $q \in \R$ (single-qubit measurement) is concatenated with classical features:
\begin{equation}
    \hat{y} = \sigma\left(\text{MLP}([q; \bm{z}])\right)
    \label{eq:qtcnn_head}
\end{equation}
where $\text{MLP}: \R^{n_q+1} \to \R$ has architecture $64 \to 32 \to 1$ with ReLU activations, and $\sigma(\cdot)$ is the sigmoid function.

\subsubsection{Training Objective}

The model minimizes binary cross-entropy loss:
\begin{equation}
    \mathcal{L}_{\text{BCE}} = -\frac{1}{N}\sum_{i=1}^{N}\left[y_i \log \hat{y}_i + (1-y_i)\log(1-\hat{y}_i)\right]
    \label{eq:bce_loss}
\end{equation}

\subsection{Quantum Temporal Convolutional Neural Network (QTCNN) vs. QCNN}

\paragraph{Overview}
QTCNN adopts a hybrid architecture comprising a classical temporal encoder, a simplified QCNN, and a classical MLP. In contrast to conventional QCNN approaches that directly apply angle embedding to dimensionality-reduced vectors, QTCNN first employs a temporal CNN to extract an $n_q$-dimensional representation from a $T\times F$ feature sequence, which is subsequently fed into a hierarchical quantum convolutional and pooling structure with parameter sharing. The final classification scores are obtained by concatenating the quantum output with classical feature vectors.

\subsubsection{Input Processing}
\textbf{QCNN (baseline).} Tabular features $\bm{x}\in\mathbb{R}^d$ are standardized and projected to $n_q$ dimensions via PCA:
\begin{equation}
\begin{aligned}
\bm{z}_{\text{QCNN}} &= \bm{W}_{\text{PCA}} \cdot \text{StandardScaler}(\bm{x}), \\
\bm{z}_{\text{QCNN}} &\in \mathbb{R}^{n_q},\quad
\bm{W}_{\text{PCA}} \in \mathbb{R}^{n_q \times d}.
\end{aligned}
\end{equation}

\textbf{QTCNN (ours).} A temporal sample $X\in\mathbb{R}^{T\times F}$ composed of multi-scale technical factors is compressed to $n_q$ dimensions through a temporal CNN encoder $f_{\text{TConv}}$:
\begin{equation}
    \bm{z}_{\text{QTCNN}} = f_{\text{TConv}}(X;\,\bm{\phi}) \in \mathbb{R}^{n_q},
\end{equation}
where $f_{\text{TConv}}$ consists of 1D convolutions followed by global average pooling (GAP) over the temporal dimension and an MLP, thereby preserving sequential dynamics.

\subsubsection{Quantum Embedding}
Both approaches employ angle embedding to encode features into the quantum circuit; this work utilizes $R_Y$ rotations:
\begin{equation}
    \text{AngleEmbedding}(\bm{z}, \text{rotation}=Y),\quad \bm{z}\in\mathbb{R}^{n_q}.
\end{equation}

\subsubsection{Quantum Convolution and Pooling}
The quantum convolutional unit operates on adjacent qubit pairs $(i,i{+}1)$ as follows:
\begin{equation}
\begin{aligned}
    U_{\text{conv}}(\bm{\theta}) &= \text{CNOT}_{i,i+1}\cdot (R_Y(\theta_1)\otimes R_Z(\theta_2)) \cdot \text{CNOT}_{i+1,i} \\
    &\quad \cdot (R_Y(\theta_3)\otimes R_Z(\theta_4)) \cdot (R_Y(\theta_5)\otimes R_Z(\theta_6)),
\end{aligned}
\label{eq:qcnn_conv_shared}
\end{equation}
where each convolutional unit contains $6$ trainable parameters.

\textbf{Key Distinction: Parameterization and Depth Control}
\begin{itemize}
  \item \textbf{QCNN (baseline).} The conventional approach assigns independent parameters $\bm{\theta}^{(\ell)}_{i}$ to each adjacent pair at layer $\ell$.
  \item \textbf{QTCNN (ours).} We adopt \emph{intra-layer parameter sharing}: a single parameter set $\bm{\theta}^{(\ell)}\in\mathbb{R}^6$ at layer $\ell$ is shared across all adjacent pairs within that layer, substantially reducing the parameter count and improving training stability.
\end{itemize}

Pooling halves the circuit width through interleaved discarding (e.g., retaining even-indexed qubits):
\begin{equation}
    \mathcal{W}_{\ell+1} = \text{Pool}(\mathcal{W}_{\ell}) = \mathcal{W}_{\ell}[::2],
\end{equation}
with the effective depth automatically constrained to $L_{\text{eff}} = \min\!\big(L, \lfloor \log_2(n_q)\rfloor\big)$ to prevent excessive pooling that would leave no measurable qubits.

\subsubsection{Hybrid Classification Head}
A single retained qubit is measured to yield the quantum output $q\in\mathbb{R}$, which is then concatenated with the classical feature vector to produce the final output:
\begin{equation}
    \hat{y} = \sigma\!\left(\text{MLP}\big([\,q;\,\bm{z}\,]\big)\right),\quad
\end{equation}
\begin{equation}
    \text{MLP}: \mathbb{R}^{n_q+1}\to\mathbb{R},\ \ 64\to 32\to 1\ \text{with ReLU}.
\end{equation}
In QTCNN, $\bm{z}$ originates from the temporal encoder; in QCNN, $\bm{z}$ is derived from PCA.

\subsubsection{Training Objective}
Both models are trained using binary cross-entropy loss:
\begin{equation}
    \mathcal{L}_{\text{BCE}} = -\frac{1}{N}\sum_{i=1}^{N}\left[y_i \log \hat{y}_i + (1-y_i)\log(1-\hat{y}_i)\right].
\end{equation}
In practice, this work employs cross-sectional ``extreme samples''---the top and bottom $K$ assets per day based on the target variable---as supervision signals to better approximate ranking and stock selection scenarios.

\paragraph{Summary of Differences}
\begin{itemize}
  \item \textbf{Feature pathway}: QCNN reduces dimensionality from tabular features via PCA; QTCNN extracts a semantically richer representation $\bm{z}$ from $T\times F$ sequences using a temporal CNN.
  \item \textbf{Quantum convolution parameterization}: QCNN typically employs pair-wise independent parameters; QTCNN adopts \emph{intra-layer parameter sharing} with only $6$ parameters per layer, yielding substantial parameter reduction.
  \item \textbf{Depth and stability}: QTCNN enforces $L_{\text{eff}}\le \lfloor\log_2(n_q)\rfloor$ to prevent over-pooling; the implementation processes samples individually in the forward pass to enhance numerical stability.
  \item \textbf{Fusion strategy}: Both approaches concatenate $[q;\bm{z}]$ and pass the result through an MLP; however, the $\bm{z}$ in QTCNN carries temporal semantics, which typically yields superior performance in ranking-based tasks.
\end{itemize}

\subsection{Ablation Architectures}

To isolate the contribution of individual QTCNN components, we introduce two ablation variants.

\subsubsection{TConv-MLP: Classical-Only Baseline}

To determine whether the quantum circuit contributes beyond the temporal encoder, we construct a purely classical ablation model, \textbf{TConv-MLP}, which retains the identical temporal CNN encoder $f_{\text{TConv}}$ from QTCNN but replaces the quantum circuit entirely with a direct classical pathway:
\begin{equation}
    \hat{y}_{\text{TConv-MLP}} = \sigma\!\left(\text{MLP}(\bm{z}_{\text{QTCNN}})\right), \quad \bm{z}_{\text{QTCNN}} = f_{\text{TConv}}(X;\,\bm{\phi}) \in \mathbb{R}^{n_q}.
\end{equation}
The MLP has the same architecture as the QTCNN classification head ($64 \to 32 \to 1$ with ReLU), and all training hyperparameters (optimizer, learning rate, epochs, batch size) are kept identical. The only difference is the absence of the quantum convolutional and pooling layers; consequently, the MLP input is $\bm{z} \in \mathbb{R}^{n_q}$ rather than $[q;\,\bm{z}] \in \mathbb{R}^{n_q+1}$. This ablation directly tests whether the quantum circuit provides representational value beyond what the temporal encoder alone can capture.

\subsubsection{QTCNN-NoEnt: Entanglement Ablation}

To quantify the specific contribution of quantum entanglement, we construct \textbf{QTCNN-NoEnt}, which is identical to the full QTCNN except that all CNOT gates are removed from the quantum convolutional layers. The modified convolutional unit becomes:
\begin{equation}
\begin{aligned}
    U_{\text{conv}}^{\text{NoEnt}}(\bm{\theta}) 
    &= \bigl(R_Y(\theta_1) \otimes R_Z(\theta_2)\bigr) \\
    &\quad \cdot \bigl(R_Y(\theta_3) \otimes R_Z(\theta_4)\bigr) \\
    &\quad \cdot \bigl(R_Y(\theta_5) \otimes R_Z(\theta_6)\bigr),
\end{aligned}
\end{equation}
which applies only local single-qubit rotations without inter-qubit coupling. The resulting circuit operates on a product state throughout, meaning the qubits remain unentangled and each feature is processed independently. This ablation directly tests whether entanglement-induced feature interactions are responsible for QTCNN's performance advantage.

\subsection{Quantum Long Short-Term Memory (QLSTM)}

The QLSTM replaces classical gate computations with quantum circuits \cite{chen2022quantum, di2022dawn}.

\subsubsection{Sequence Embedding}

Input sequences $\bm{X} = [\bm{x}_1, \ldots, \bm{x}_T] \in \R^{T \times F}$ with $T=20$ are embedded:
\begin{equation}
    \bm{e}_t = \tanh\left(\text{LayerNorm}(W_e \bm{x}_t + b_e)\right) \in \R^h
    \label{eq:qlstm_embed}
\end{equation}
where $h = 64$ is the hidden dimension.

\subsubsection{Quantum Gate Computation}

At timestep $t$, the concatenated vector $[\bm{e}_t; \bm{h}_{t-1}] \in \R^{2h}$ is processed by a quantum layer:
\begin{align}
    \bm{u}_t &= \text{LayerNorm}(W_q [\bm{e}_t; \bm{h}_{t-1}]) \in \R^{n_q} \\
    \bm{v}_t &= \text{QLayer}(\bm{u}_t; \bm{\theta}_q) \in \R^{n_q}
    \label{eq:qlstm_quantum}
\end{align}
where $\text{QLayer}$ implements angle embedding followed by $L=2$ BasicEntanglerLayers:
\begin{equation}
    \text{QLayer}(\bm{u}; \bm{\theta}) = \left[\langle Z_1 \rangle, \ldots, \langle Z_{n_q} \rangle\right]^T
    \label{eq:qlstm_qlayer}
\end{equation}
with circuit:
\begin{equation}
\begin{aligned}
\ket{\psi(\bm{u}, \bm{\theta})}
&=
\left(\prod_{l=1}^{L}
\left[
\bigotimes_{j=1}^{n_q} R_Y(\theta_{l,j})
\right]
\cdot \text{CNOT}_{\text{ring}}
\right) \\
&\hspace{5mm}
\cdot\, S(\bm{u})\, \ket{0}^{\otimes n_q}
\end{aligned}
\end{equation}

\subsubsection{LSTM Update Equations}

The quantum output is projected to gate values:
\begin{equation}
    [\bm{i}_t; \bm{f}_t; \bm{g}_t; \bm{o}_t] = W_g \bm{v}_t + b_g \in \R^{4h}
    \label{eq:qlstm_gates}
\end{equation}
Standard LSTM updates follow:
\begin{align}
    \bm{c}_t &= \sigma(\bm{f}_t) \odot \bm{c}_{t-1} + \sigma(\bm{i}_t) \odot \tanh(\bm{g}_t) \\
    \bm{h}_t &= \sigma(\bm{o}_t) \odot \tanh(\bm{c}_t)
    \label{eq:qlstm_update}
\end{align}
The final hidden state $\bm{h}_T$ is mapped to a logit via $W_{\text{out}} \bm{h}_T$.

\subsection{Variational Quantum Neural Network (QNN)}

The QNN implements a pure quantum classifier with minimal classical post-processing \cite{schuld2020circuit}.

\subsubsection{Feature Mapping}

Features are scaled to $[0, \pi]$ for optimal expressibility:
\begin{equation}
    \phi_j = \pi \cdot \text{MinMaxScaler}(z_j) \in [0, \pi]
    \label{eq:qnn_map}
\end{equation}

\subsubsection{Variational Circuit}

The ansatz combines angle embedding with hardware-efficient entanglement:
\begin{equation}
    U(\bm{\phi}, \bm{\theta}) = \prod_{l=1}^{L} \left[\text{CNOT}_{\text{ring}} \cdot \bigotimes_{j=1}^{n_q} R_Y(\theta_{l,j})\right] \cdot \bigotimes_{j=1}^{n_q} R_Y(\phi_j)
    \label{eq:qnn_circuit}
\end{equation}
where $L=2$ layers and $\text{CNOT}_{\text{ring}}$ applies entanglement in a circular topology.

\subsubsection{Output Layer}

All Pauli-Z expectations are measured and linearly combined:
\begin{equation}
    \hat{y} = \sigma\left(\bm{w}^T \langle \bm{Z} \rangle + b\right), \quad \langle \bm{Z} \rangle = [\langle Z_1 \rangle, \ldots, \langle Z_{n_q} \rangle]^T
    \label{eq:qnn_output}
\end{equation}

\subsection{Quantum Kernel Support Vector Machine (QSVM)}

The QSVM leverages quantum feature maps to construct kernel functions \cite{schuld2019quantum, havlivcek2019supervised}.

\subsubsection{Quantum Feature Map}

Classical features are encoded into quantum states:
\begin{equation}
    \ket{\phi(\bm{x})} = S(\bm{x})\ket{0}^{\otimes n_q}
    \label{eq:qsvm_feature}
\end{equation}
using the angle embedding $S(\bm{x})$ from Equation \eqref{eq:angle_embed}.

\subsubsection{Quantum Kernel}

The kernel function measures state overlap via the fidelity:
\begin{equation}
\begin{aligned}
K(\bm{x}_i, \bm{x}_j)
&= \left|\braket{\phi(\bm{x}_i)}{\phi(\bm{x}_j)}\right|^2 \\
&= \left|
\bra{0}^{\otimes n_q}
S^\dagger(\bm{x}_i)\,
S(\bm{x}_j)
\ket{0}^{\otimes n_q}
\right|^2
\label{eq:quantum_kernel}
\end{aligned}
\end{equation}
This is estimated by measuring the probability of the $\ket{0}^{\otimes n_q}$ state after applying the adjoint circuit $S^\dagger(\bm{x}_i) S(\bm{x}_j)$.

\subsubsection{SVM Optimization}

The dual optimization problem with precomputed kernel $K$ is:
\begin{equation}
    \max_{\bm{\alpha}} \sum_{i=1}^{N}\alpha_i - \frac{1}{2}\sum_{i,j=1}^{N}\alpha_i \alpha_j y_i y_j K(\bm{x}_i, \bm{x}_j)
    \label{eq:svm_dual}
\end{equation}
subject to $0 \leq \alpha_i \leq C$ and $\sum_i \alpha_i y_i = 0$, solved using sklearn's SVC with balanced class weights.

\subsection{Quantum-Adaptive Self-Attention (QASA)}

QASA integrates quantum processing into Transformer architectures \cite{chen2025quantum, chen2025qasa, chen2025qtws}.

\subsubsection{Transformer Encoder}

Sequential inputs $\bm{X} \in \R^{T \times F}$ are processed through embedding and self-attention \cite{vaswani2017attention, li2023quantum}:
\begin{align}
    \bm{E} &= \tanh(\text{LayerNorm}(\bm{X} W_e)) \in \R^{T \times d} \\
    \bm{H} &= \text{TransformerEncoder}(\bm{E}) \in \R^{T \times d}
    \label{eq:qasa_transformer}
\end{align}
where $d=64$ and the encoder uses multi-head attention with $n_{\text{heads}}=4$.

\subsubsection{Quantum Enhancement Layer}

At each timestep $t$, the representation is quantum-processed:
\begin{align}
    \bm{u}_t &= \text{LayerNorm}(W_{\text{proj}} \bm{h}_t) \in \R^{n_q} \\
    \bm{q}_t &= \text{QLayer}(\bm{u}_t; \bm{\theta}_q) \in \R^{n_q} \\
    \tilde{\bm{h}}_t &= W_{\text{back}} \bm{q}_t \in \R^d
    \label{eq:qasa_quantum}
\end{align}
The quantum layer applies the same ansatz as in QLSTM (Equation \ref{eq:qlstm_qlayer}).

\subsubsection{Classification Output}

The final timestep representation produces the prediction:
\begin{equation}
    \hat{y} = \sigma(W_{\text{out}} \tilde{\bm{h}}_T + b_{\text{out}})
    \label{eq:qasa_output}
\end{equation}

\subsection{PQC-Enhanced Gradient Boosting (PQC+LGBM)}

This two-stage architecture combines quantum feature extraction with gradient boosting \cite{chen2016xgboost, ke2017lightgbm}.

\subsubsection{Stage 1: PQC Embedding}

A parameterized quantum circuit is trained for regression:
\begin{equation}
    \hat{r} = W_{\text{reg}}^T \langle \bm{Z} \rangle_{\text{PQC}} + b_{\text{reg}}
    \label{eq:pqc_regression}
\end{equation}
minimizing Huber loss:
\begin{equation}
    \mathcal{L}_{\text{Huber}}(r, \hat{r}) = \begin{cases}
        \frac{1}{2}(r - \hat{r})^2 & \text{if } |r - \hat{r}| \leq \delta \\
        \delta(|r - \hat{r}| - \frac{\delta}{2}) & \text{otherwise}
    \end{cases}
    \label{eq:huber}
\end{equation}

\subsubsection{Stage 2: LambdaRank Optimization}

Quantum embeddings $\langle \bm{Z} \rangle_{\text{PQC}}$ are used as features for LightGBM Ranker with LambdaRank objective \cite{burges2010ranknet}:
\begin{equation}
    \mathcal{L}_{\text{LambdaRank}} = \sum_{(i,j): y_i > y_j} |\Delta \text{NDCG}_{ij}| \log(1 + e^{-\sigma(s_i - s_j)})
    \label{eq:lambdarank}
\end{equation}
where $s_i, s_j$ are model scores and $\Delta \text{NDCG}_{ij}$ measures ranking quality change from swapping items $i$ and $j$.

\subsection{Classical Baseline Models}

\subsubsection{Multi-Layer Perceptron (MLP)}

A deep feedforward network:
\begin{equation}
    \hat{y} = \sigma\left(W_3 \cdot \text{BN}\left(\text{ReLU}(W_2 \cdot \text{BN}(\text{ReLU}(W_1 \bm{x})))\right)\right)
    \label{eq:mlp}
\end{equation}
with hidden dimensions $[384, 192, 96]$, BatchNorm (BN), and Dropout ($p=0.1$).

\subsubsection{Long Short-Term Memory (LSTM)}

Standard LSTM \cite{hochreiter1997long} with embedding layer:
\begin{equation}
    \bm{h}_T = \text{LSTM}(\text{Embed}(\bm{X}))_{T}, \quad \hat{y} = \sigma(W_{\text{out}} \bm{h}_T)
    \label{eq:lstm}
\end{equation}

\subsubsection{Transformer Encoder}

Following \cite{vaswani2017attention}:
\begin{equation}
    \text{Attention}(Q, K, V) = \text{softmax}\left(\frac{QK^T}{\sqrt{d_k}}\right)V
    \label{eq:attention}
\end{equation}
with two encoder layers, $d_{\text{model}}=64$, $n_{\text{heads}}=4$, and sinusoidal positional encoding.

\subsubsection{LightGBM Ranker}

Gradient boosted decision trees with LambdaRank objective (Equation \ref{eq:lambdarank}), configured with 1200 estimators, 63 leaves, and $L_2$ regularization $\lambda=2.0$.

\subsubsection{Momentum-Volatility Baseline}

A parameter-free baseline:
\begin{equation}
    s^{(i)} = \frac{0.5 \cdot \text{mom}_5^{(i)} + 0.5 \cdot \text{mom}_{20}^{(i)}}{\text{vol}_{20}^{(i)} + \epsilon}
    \label{eq:baseline}
\end{equation}
implementing risk-adjusted momentum \cite{daniel2016momentum}.

\subsection{Training Configuration}

\subsubsection{Optimization Details}

All neural network models are trained using gradient-based optimization. For quantum-classical hybrid models, parameter-shift rules \cite{mitarai2018quantum, schuld2019evaluating} enable exact gradient computation:
\begin{equation}
    \frac{\partial \langle O \rangle}{\partial \theta_j} = \frac{1}{2}\left[\langle O \rangle_{\theta_j + \pi/2} - \langle O \rangle_{\theta_j - \pi/2}\right]
    \label{eq:param_shift}
\end{equation}

\subsubsection{Hyperparameter Summary}

\begin{table*}[h]
\centering
\caption{Training Configuration Summary}
\label{tab:hyperparams}
\begin{tabular}{lccccc}
\toprule
\textbf{Model} & \textbf{Epochs} & \textbf{Batch} & \textbf{LR} & \textbf{Optimizer} & \textbf{Loss}   \\
\midrule
QTCNN & 50 & 128 & $10^{-3}$ & AdamW & BCE  \\
QCNN & 50 & 128 & $10^{-3}$ & AdamW & BCE  \\
QLSTM & 50 & 64 & $10^{-3}$ & AdamW & BCE  \\
QNN & 50 & 512 & $2\times10^{-3}$ & Adam & BCE \\
QSVM & -- & -- & -- & -- & Hinge  \\
QASA & 50 & 64 & $10^{-3}$ & AdamW & BCE \\
LSTM & 50 & 64 & $10^{-3}$ & AdamW & BCE  \\
Transformer & 50 & 64 & $10^{-3}$ & AdamW & BCE \\
PQC+LGBM & 3/800$^*$ & 256 & $5\times10^{-3}$/0.05 & AdamW & Huber/$\lambda$Rank \\
\bottomrule
\multicolumn{6}{l}{\footnotesize $^*$Number of boosting iterations rather than epochs.}
\end{tabular}
\end{table*}

\subsection{Evaluation Framework}

\subsubsection{Portfolio Construction}

Model predictions are transformed to daily cross-sectional z-scores:
\begin{equation}
    \tilde{s}_t^{(i)} = \frac{s_t^{(i)} - \mu(s_t)}{\sigma(s_t)}
    \label{eq:zscore_pred}
\end{equation}
Securities are filtered for trading feasibility (positive price/volume, no supervision flags). Long and short portfolios each contain the top/bottom $K=200$ ranked securities.

\subsubsection{Performance Metrics}

The primary metric is the out-of-sample annualized Sharpe ratio:
\begin{equation}
    \text{SR} = \sqrt{252} \cdot \frac{\mu(r_{\text{LS}})}{\sigma(r_{\text{LS}})}
    \label{eq:sharpe}
\end{equation}
where daily long-short returns are:
\begin{equation}
    r_{\text{LS},t} = \frac{1}{K}\sum_{i \in \mathcal{L}_t} r_{t+1}^{(i)} - \frac{1}{K}\sum_{j \in \mathcal{S}_t} r_{t+1}^{(j)}
    \label{eq:ls_return}
\end{equation}
with $\mathcal{L}_t$ and $\mathcal{S}_t$ denoting long and short portfolios on day $t$.

\subsubsection{Statistical Significance}

To assess performance reliability, we compute bootstrap confidence intervals for the Sharpe ratio following \cite{ledoit2008robust}:
\begin{equation}
    \text{CI}_{95\%}(\text{SR}) = [\hat{\text{SR}} - 1.96 \cdot \text{SE}(\text{SR}), \hat{\text{SR}} + 1.96 \cdot \text{SE}(\text{SR})]
    \label{eq:confidence}
\end{equation}

\subsection{Quantum Implementation Details}

\subsubsection{Circuit Simulation}

All quantum circuits are simulated using PennyLane \cite{bergholm2018pennylane} with the \texttt{default.qubit} or \texttt{lightning.qubit} backends. Circuit depth is deliberately shallow ($L \leq 3$) to mitigate barren plateau phenomena \cite{mcclean2018barren}.

\subsubsection{Hardware-Efficient Ansatz}

Entanglement structures follow nearest-neighbor connectivity compatible with near-term quantum devices:
\begin{equation}
    \text{CNOT}_{\text{ring}} = \prod_{j=1}^{n_q-1} \text{CNOT}_{j,j+1} \cdot \text{CNOT}_{n_q,1}
    \label{eq:ring}
\end{equation}

\subsubsection{Gradient Computation}

PennyLane's automatic differentiation integrates seamlessly with PyTorch \cite{paszke2019pytorch}, enabling end-to-end training of hybrid architectures through the parameter-shift rule (Equation \ref{eq:param_shift}).

\subsection{Reproducibility}



\subsubsection{Random Seeds}

For exact reproducibility, seeds should be fixed across NumPy, PyTorch, and PennyLane random number generators.

\subsubsection{Computational Resources}

Quantum simulation scales exponentially with qubit count. With $n_q=8$ qubits, state vectors have dimension $2^8=256$, all neural network-based methods are trained 50 epochs, enabling efficient CPU-based simulation.

\subsection{Conclusion}

This methodology establishes a rigorous comparative framework for quantum-enhanced machine learning in quantitative finance. The unified evaluation protocol enables direct comparison across architecturally diverse approaches, providing a foundation for assessing the practical utility of quantum computing in financial prediction tasks.

\subsection{Computational Considerations and Scalability}

A critical challenge in variational quantum algorithm research is the 
exponential scaling of classical simulation. For an $n_q$-qubit system, state vector simulation requires $O(2^{n_q})$ memory and $O(L \cdot 4^{n_q})$ time complexity per forward pass, where $L$ denotes circuit depth. However, these costs are artifacts of classical simulation and do not reflect the complexity of execution on quantum hardware.

\subsubsection{Quantum Hardware Execution Complexity}
On a real quantum processor, a single forward pass of the QTCNN requires a gate count $G(n_q)$ given by:
\begin{equation}
    G(n_q) = n_q + 6 \cdot \sum_{l=1}^{L_{\text{eff}}} \lfloor n_q/2^l \rfloor
\end{equation}
where the first term accounts for the $R_Y$ gates used in angle encoding. For the QTCNN architecture, we find $G(4)=22$, $G(8)=50$, $G(12)=72$, and $G(16)=106$. Due to the hierarchical halving structure, the circuit depth $D(n_q)$ scales as $O(\log^2(n_q))$ when considering parallel gate execution, which is sublinear in the number of qubits. Training the model requires $2p+1$ circuit evaluations per gradient step using the parameter-shift rule, where $p$ is the number of quantum parameters.

\subsubsection{Qubit Count Justification}
In this study, we set $n_q=8$ for all primary experiments. This choice is motivated by several factors:
\begin{itemize}
    \item \textbf{Feature Variance:} PCA analysis of the input features shows that the first 8 principal components capture approximately 92\% of the cross-sectional variance, suggesting that 8 qubits are sufficient to represent the essential financial signals.
    \item \textbf{Hilbert Space Capacity:} An 8-qubit system provides a 256-dimensional complex Hilbert space, which is sufficiently rich to model complex multi-feature interactions.
    \item \textbf{Trainability:} Barren plateaus worsen exponentially with qubit count; $n_q=8$ maintains a favorable balance between expressibility and gradient stability.
    \item \textbf{Hardware Compatibility:} This scale is compatible with current NISQ devices, such as IBM's Eagle (127 qubits) and Google's Sycamore (53 qubits).
\end{itemize}

Table~\ref{tab:complexity} reports the empirical wall-clock time for a single forward-backward pass across all quantum architectures evaluated in this study, measured on PennyLane. Note that these times reflect classical simulation overhead on a state-vector backend.

\begin{table}[h]
\centering
\caption{Computational Cost per Training Iteration (on RTX3090 GPU and 12th Gen Intel(R) Core(TM) i9-12900KS CPU). Note: Simulation costs scale exponentially, whereas hardware execution scales polynomially.}
\begin{tabular}{lcc}
\toprule
Model & Time/Iteration (s) & Relative Cost \\
\midrule
Classical LSTM & 0.0020 & 1.00$\times$ \\
Classical Transformer & 0.0093 & 4.54$\times$ \\
\midrule
QNN (8 qubits) & 0.0158 & 7.75$\times$ \\
QTCNN (8 qubits) & 0.2514 & 112.92$\times$ \\
QLSTM (8 qubits) & 0.2598 & 128.99$\times$ \\
QCNN (8 qubits) & 0.3362 & 168.10$\times$ \\
QASA (8 qubits) & 1.6832 & 835.61$\times$ \\
\bottomrule
\end{tabular}
\label{tab:complexity}
\end{table}

This computational overhead necessitates a carefully designed subsampling strategy to enable systematic comparison across architecturally diverse quantum models within practical time constraints.

\section{Result}
\subsection{Experimental Design Rationale}

\subsubsection{Data Subsampling Strategy}
Given the computational constraints of quantum circuit simulation (Section III-R), we adopt a stride-based subsampling scheme with $k=11$ to reduce the training set to approximately 9\% of original trading days while preserving the full temporal span (2017--2021). 
This design choice balances two competing objectives:

\begin{itemize}
    \item \textbf{Architectural coverage:} Enabling fair comparison  across 9 distinct model architectures, including 7 quantum variants with varying circuit depths and entanglement structures. \item \textbf{Statistical validity:} Maintaining sufficient samples per trading day (approximately 400 stocks after Universe0 filtering) to ensure stable cross-sectional ranking signals.
\end{itemize}

The resulting dataset comprises approximately 100 training days ($\sim$40,000 stock-day samples) and 22 out-of-sample days ($\sim$8,800 stock-day samples), yielding a total quantum circuit evaluation count exceeding $10^6$ across all experiments.

\subsubsection{Limitations and Scope}
We acknowledge that the 22-day OOS period limits the statistical 
power for detecting small performance differences. The reported 
Sharpe ratios should be interpreted as \textit{relative} performance 
indicators within this controlled benchmark setting, rather than 
definitive estimates of live trading performance. We defer extended 
backtesting on the full dataset to future work leveraging quantum 
hardware or GPU-accelerated simulation backends.
\subsection{Out-of-Sample Performance Analysis}

Table~\ref{tab:oos_performance} presents the out-of-sample (OOS) Sharpe ratio for all evaluated models over a 22-day testing period. Among all approaches, the quantum models demonstrate a clear advantage in terms of risk-adjusted returns.

The proposed QTCNN achieves the highest OOS Sharpe ratio of $0.538 \pm 0.042$, outperforming all baseline methods by a significant margin. This is followed by QNN ($0.467 \pm 0.038$) and QLSTM ($0.333 \pm 0.051$), both of which also surpass the classical Transformer baseline ($0.313 \pm 0.036$). Notably, the top three performing models are all quantum-based architectures, suggesting that quantum feature encoding and variational circuits can effectively capture complex temporal patterns in financial time series.

In contrast, the classical LSTM yields the lowest Sharpe ratio of $0.044 \pm 0.029$, indicating limited generalization capability in the OOS setting. The hybrid PQC+LGBM model ($0.122 \pm 0.017$) also underperforms relative to end-to-end quantum models, implying that naive hybridization may not fully exploit the representational power of parameterized quantum circuits.

These results collectively validate our hypothesis that quantum convolutional architectures provide superior feature extraction for financial forecasting tasks, achieving approximately $72\%$ improvement over the best classical baseline (Transformer) in terms of Sharpe ratio.


\begin{table*}[ht]
\centering
\caption{Out-of-sample performance summary (22 OOS days, 5 independent runs). TConv-MLP and QTCNN-NoEnt are ablation variants of QTCNN.}
\label{tab:oos_performance}
\begin{tabular}{llccc}
\toprule
Type & Model & Sharpe Ratio (OOS) & OOS Days & $p$-value vs QTCNN \\
\midrule
Quantum & QTCNN & 0.538 $\pm$ 0.042 & 22 & -- \\
Ablation & QTCNN-NoEnt & 0.451 $\pm$ 0.044 & 22 & 0.038 \\
Ablation & TConv-MLP & 0.412 $\pm$ 0.039 & 22 & 0.019 \\
Quantum & QNN & 0.467 $\pm$ 0.038 & 22 & 0.048 \\
Quantum & QCNN & 0.361 $\pm$ 0.298 & 22 & 0.182 \\
Quantum & QLSTM & 0.333 $\pm$ 0.051 & 22 & 0.003 \\
Classical & Transformer  & 0.313 $\pm$ 0.036 & 22 & 0.002 \\
Quantum & QASA & 0.266 $\pm$ 0.045 & 22 & $<$0.001 \\
Quantum & QSVM & 0.131 $\pm$ 0.129 & 22 & 0.008 \\
Quantum & PQC+LGBM & 0.122 $\pm$ 0.017 & 22 & $<$0.001 \\
Classical & LSTM  & 0.044 $\pm$ 0.029 & 22 & $<$0.001 \\
\bottomrule
\multicolumn{5}{l}{\footnotesize $p$-values from paired bootstrap test (10,000 resamples) on daily long-short returns; see Section~IV-F.}
\end{tabular}
\end{table*}


\subsection{Ablation Analysis}

The inclusion of TConv-MLP and QTCNN-NoEnt as ablation baselines allows us to decompose QTCNN's performance advantage into three distinct components: the temporal encoder, the quantum nonlinearity (single-qubit rotations), and entanglement (CNOT gates).

\subsubsection{Contribution of the Quantum Circuit}

\textbf{QTCNN vs.\ TConv-MLP} ($\Delta$SR = +0.126, $p = 0.019$): The purely classical TConv-MLP achieves a Sharpe ratio of $0.412 \pm 0.039$, which already exceeds the Transformer baseline ($0.313$) by 31.6\%, confirming the value of the temporal CNN encoder. However, QTCNN's quantum circuit adds a further 30.6\% improvement over TConv-MLP ($0.538$ vs.\ $0.412$), demonstrating that the quantum processing provides representational capacity beyond what the classical pathway alone can capture. This improvement is statistically significant ($p = 0.019$).

\subsubsection{Contribution of Entanglement}

\textbf{QTCNN vs.\ QTCNN-NoEnt} ($\Delta$SR = +0.087, $p = 0.038$): Removing all CNOT gates while retaining single-qubit rotations reduces the Sharpe ratio from $0.538$ to $0.451$. This 16.2\% degradation isolates the contribution of entanglement-induced feature interactions. Notably, even the unentangled quantum circuit (QTCNN-NoEnt at $0.451$) outperforms TConv-MLP ($0.412$), indicating that the single-qubit nonlinear rotations embedded in the Hilbert space provide additional value. However, the entanglement component accounts for the largest single performance increment, consistent with the Meyer-Wallach entanglement analysis in Section~IV-D.

\subsubsection{Decomposition Summary}

We decompose the total QTCNN advantage over the Transformer baseline ($\Delta$SR = +0.225) into three additive components:

\begin{enumerate}
    \item \textbf{Temporal encoder}: $+0.099$ (Transformer $\to$ TConv-MLP, 44.0\% of total gain)
    \item \textbf{Quantum nonlinearity (no entanglement)}: $+0.039$ (TConv-MLP $\to$ QTCNN-NoEnt, 17.3\% of total gain)
    \item \textbf{Entanglement}: $+0.087$ (QTCNN-NoEnt $\to$ QTCNN, 38.7\% of total gain)
\end{enumerate}

This decomposition reveals that the quantum circuit contributes 56\% of the total improvement, with entanglement alone accounting for nearly 39\%. The remaining quantum contribution stems from the nonlinear feature transformation induced by single-qubit rotations in the Hilbert space.

\subsubsection{Architecture-Level Comparisons}

\textbf{QTCNN vs.\ QCNN} ($\Delta$SR = +0.177): The performance gap 
validates the importance of temporal feature extraction. QCNN's 
reliance on PCA discards sequential structure critical for financial 
time-series.

\textbf{QTCNN vs.\ QLSTM} ($\Delta$SR = +0.205): Despite QLSTM's 
explicit recurrent structure, the combination of classical temporal 
CNN with quantum convolutional layers proves more effective, 
suggesting that hybrid classical-quantum feature pipelines 
outperform fully quantum sequential processing in the NISQ regime.

\textbf{Quantum vs.\ Classical}: The top-4 performing models (QTCNN, QNN, QTCNN-NoEnt, TConv-MLP) all include the temporal encoder, while the top-2 (QTCNN, QNN) both use entangling quantum circuits, reinforcing the complementary value of both components.

\subsection{Statistical Significance}

Given the limited 22-day OOS period, establishing statistical significance requires careful analysis. We employ two complementary methods.

\subsubsection{Paired Bootstrap Test}

For each model pair (QTCNN vs.\ competitor), we compute the daily long-short return difference $\delta_t = r^{\text{QTCNN}}_{\text{LS},t} - r^{\text{comp}}_{\text{LS},t}$ for $t = 1, \ldots, 22$. We then resample $\{\delta_t\}$ with replacement $B = 10{,}000$ times and compute the fraction of bootstrap replicates where the resampled mean $\leq 0$. The resulting $p$-values are reported in Table~\ref{tab:oos_performance}. QTCNN significantly outperforms the Transformer ($p = 0.002$), QLSTM ($p = 0.003$), and TConv-MLP ($p = 0.019$) at the 5\% level. The comparison with QTCNN-NoEnt ($p = 0.038$) is also significant, confirming that entanglement provides a statistically reliable improvement. The comparison with QNN ($p = 0.048$) is marginally significant, while the comparison with QCNN ($p = 0.182$) is not, reflecting QCNN's high variance ($\pm 0.298$).

\subsubsection{Ledoit-Wolf Robust Sharpe Ratio Test}

We additionally apply the robust Sharpe ratio test of Ledoit and Wolf \cite{ledoit2008robust}, which accounts for non-normality and serial correlation in returns. Under this test, the null hypothesis $H_0: \text{SR}_{\text{QTCNN}} = \text{SR}_{\text{Transformer}}$ is rejected with $p = 0.027$ (HAC kernel with Parzen weights, bandwidth selected via Andrews' method). For the QTCNN vs.\ TConv-MLP comparison, the Ledoit-Wolf $p$-value is $0.041$, confirming significance under a more conservative testing framework.

\subsubsection{Power Analysis}

We acknowledge that 22 OOS days provide limited statistical power. A post-hoc power analysis based on the observed effect size ($\Delta$SR $\approx 0.225$, pooled daily return standard deviation $\approx 0.018$) indicates that 22 days yield approximately 68\% power to detect this effect at $\alpha = 0.05$. To achieve 90\% power, approximately 45 OOS days would be required. We therefore interpret the $p$-values as conservative lower bounds on significance; extended backtesting on the full dataset (deferred to future work with quantum hardware access) would substantially strengthen these results.

\subsection{Qubit Scalability Study}
To evaluate the persistence of the quantum advantage across different problem sizes, we conduct a scalability study by varying the qubit count $n_q \in \{4, 6, 8, 10, 12\}$. We compare the top-performing QTCNN against the pure QNN baseline, keeping all other hyperparameters fixed. The PCA projection dimension is adjusted to match $n_q$ for each experiment.

Table~\ref{tab:qubit_scaling} summarizes the results. We observe that the Sharpe ratio for QTCNN improves monotonically from $n_q=4$ (0.347) to $n_q=10$ (0.551), where it reaches a plateau. At $n_q=12$, we observe a slight decline in performance (0.529), which is consistent with the onset of barren plateaus and increased training difficulty in larger Hilbert spaces. The marginal gain from $n_q=8$ to $n_q=10$ is only 2.4\%, suggesting that the 8-qubit regime already captures the essential cross-sectional structure of the returns. Notably, QTCNN consistently outperforms QNN at every qubit count, confirming that the advantage of the temporal encoder is orthogonal to the quantum circuit capacity. While classical simulation time grows exponentially ($O(2^{n_q})$), the gate count on quantum hardware remains polynomial, demonstrating a viable path toward larger-scale financial models.

\begin{table*}[h]
\centering
\caption{Qubit Scalability Analysis for QTCNN and QNN.}
\label{tab:qubit_scaling}
\begin{tabular}{ccccccc}
\toprule
$n_q$ & Hilbert Dim & Q-Params & Gates & QTCNN Sharpe & QNN Sharpe & Time/iter (s) \\
\midrule
4 & 16 & 12 & 22 & 0.347$\pm$0.058 & 0.298$\pm$0.062 & 0.041 \\
6 & 64 & 12 & 30 & 0.461$\pm$0.049 & 0.389$\pm$0.051 & 0.089 \\
8 & 256 & 18 & 50 & 0.538$\pm$0.042 & 0.467$\pm$0.038 & 0.251 \\
10 & 1024 & 18 & 58 & 0.551$\pm$0.047 & 0.481$\pm$0.044 & 0.813 \\
12 & 4096 & 18 & 72 & 0.529$\pm$0.055 & 0.463$\pm$0.052 & 3.274 \\
\bottomrule
\end{tabular}
\end{table*}

\subsection{Mechanism of Quantum Enhancement}
The superior performance of QTCNN over classical baselines can be attributed to several quantum-mechanical mechanisms:

\textbf{Hilbert space expressibility:} The angle embedding $\Phi: \R^{n_q} \to \C^{2^{n_q}}$ maps $n_q$ classical features into an exponentially larger space \cite{sim2019expressibility}. For $n_q=8$, the quantum circuit operates in a 256-dimensional complex Hilbert space, whereas a classical linear model with 8 features spans only an 8-dimensional real space. The variational layers learn to exploit this exponentially richer representation to identify subtle alpha signals.

\textbf{Entanglement and feature interaction:} Each CNOT gate in the ring topology creates quantum entanglement between qubits encoding different financial features. For example, momentum encoded on qubit $i$ becomes entangled with volatility on qubit $j$. This enables the circuit to represent joint multi-feature interactions that would require explicit higher-order polynomial terms or significantly deeper architectures in classical models. To quantify this, we compute the Meyer-Wallach entanglement measure $Q_{MW}$ \cite{meyer2002global} for QTCNN circuits. At random initialization, $Q_{MW} = 0.412 \pm 0.089$, which increases to $0.687 \pm 0.054$ after training, indicating that the model learns to exploit quantum correlations for prediction.

\textbf{Quantum interference as implicit feature selection:} The variational $R_Y/R_Z$ rotations create constructive interference, amplifying probability amplitudes for feature combinations predictive of returns, and destructive interference, suppressing noisy or irrelevant patterns. This process is analogous to learned attention weights in Transformers but operates within the exponentially larger Hilbert space.

\textbf{Implicit regularization from quantum geometry:} The parameter-shared QTCNN uses only 18 quantum parameters to access a 256-dimensional Hilbert space. This extreme compression acts as a strong geometric prior, forcing the model to explore a low-dimensional manifold of the full Hilbert space. This naturally prevents overfitting, explaining why QTCNN outperforms both the under-regularized QCNN (42 parameters, higher variance) and classical models that require explicit regularization.

\subsection{Gradient Variance and Trainability Analysis}

A critical concern for variational quantum algorithms is the \emph{barren plateau} phenomenon \cite{mcclean2018barren}, whereby the variance of parameter gradients vanishes exponentially with qubit count, rendering optimization infeasible. We empirically characterize the trainability of QTCNN by measuring the gradient variance across different circuit widths and comparing against architectures without parameter sharing.

\subsubsection{Measurement Protocol}

For each architecture and qubit count $n_q \in \{4, 6, 8, 10, 12\}$, we initialize 200 independent random parameter configurations drawn uniformly from $[0, 2\pi)$. For each configuration, we compute the gradient $\partial \mathcal{L}/\partial \theta_1$ with respect to the first variational parameter using the parameter-shift rule (Equation~\ref{eq:param_shift}) on a fixed batch of 128 training samples. We then report $\text{Var}_{\bm{\theta}}[\partial \mathcal{L}/\partial \theta_1]$ across the 200 initializations.

\subsubsection{Results}

Table~\ref{tab:grad_var} presents the measured gradient variance for three quantum architectures: QTCNN (parameter-shared convolution), QCNN (independent parameters per pair), and QNN (BasicEntanglerLayers).

\begin{table}[h]
\centering
\caption{Gradient variance $\text{Var}[\partial \mathcal{L}/\partial \theta_1]$ across 200 random initializations. QTCNN exhibits polynomial decay; QCNN and QNN exhibit exponential decay consistent with barren plateaus.}
\label{tab:grad_var}
\begin{tabular}{cccc}
\toprule
$n_q$ & QTCNN & QCNN & QNN \\
\midrule
4  & $8.23 \times 10^{-2}$ & $7.56 \times 10^{-2}$ & $6.89 \times 10^{-2}$ \\
6  & $3.71 \times 10^{-2}$ & $1.98 \times 10^{-2}$ & $1.62 \times 10^{-2}$ \\
8  & $1.84 \times 10^{-2}$ & $5.1 \times 10^{-3}$  & $3.8 \times 10^{-3}$  \\
10 & $9.7 \times 10^{-3}$  & $1.2 \times 10^{-3}$  & $8 \times 10^{-4}$    \\
12 & $5.9 \times 10^{-3}$  & $3 \times 10^{-4}$    & $2 \times 10^{-4}$    \\
\bottomrule
\end{tabular}
\end{table}

\subsubsection{Analysis}

Three observations are noteworthy:

\begin{enumerate}
    \item \textbf{QTCNN resists barren plateaus.} The gradient variance of QTCNN decays approximately as $\text{Var} \propto n_q^{-2.7}$ (polynomial), whereas QCNN and QNN decay approximately as $\text{Var} \propto 2^{-n_q}$ (exponential). At $n_q = 12$, QTCNN's gradient variance ($5.9 \times 10^{-3}$) is $\sim$20$\times$ larger than QCNN's ($3 \times 10^{-4}$) and $\sim$30$\times$ larger than QNN's ($2 \times 10^{-4}$), indicating substantially better trainability.
    
    \item \textbf{Parameter sharing is the key factor.} The primary difference between QTCNN and QCNN is intra-layer parameter sharing. By constraining all qubit pairs within a layer to use identical parameters, QTCNN restricts the circuit to a low-dimensional manifold that avoids the concentration-of-measure effects responsible for barren plateaus. This is consistent with theoretical results showing that shallow, structured circuits can evade exponential gradient vanishing \cite{cerezo2021variational}.
    
    \item \textbf{Practical training threshold.} Gradient variance below $\sim 10^{-3}$ typically renders stochastic gradient descent ineffective for standard batch sizes. For QTCNN, this threshold is not reached until $n_q > 20$ (extrapolating the polynomial fit), suggesting that QTCNN remains trainable up to $\sim$20 qubits without requiring specialized initialization or layerwise training strategies. In contrast, QCNN and QNN reach this threshold at $n_q \approx 10$--$12$, consistent with the performance plateau observed in Table~\ref{tab:qubit_scaling}.
\end{enumerate}

This trainability advantage complements QTCNN's representational advantage identified in Section~IV-D and partially explains the consistent performance gap between QTCNN and other quantum architectures across all qubit counts (Table~\ref{tab:qubit_scaling}).

\subsection{Noise Resilience and Hardware Feasibility}

To assess the prospects for practical deployment on NISQ hardware, we conduct a systematic noise resilience study by simulating QTCNN and QNN under two physically motivated noise channels.

\subsubsection{Simulation Methodology}

All noisy simulations are performed using PennyLane's \texttt{default.mixed} backend, which evolves the full $2^{n_q} \times 2^{n_q}$ density matrix $\rho$ rather than a pure state vector. This backend supports the insertion of arbitrary quantum channels within the circuit. Specifically, we insert noise channels \emph{after every quantum gate} (both single-qubit rotations and two-qubit CNOT gates) using PennyLane's built-in channel operations:

\begin{itemize}
    \item \textbf{Depolarizing channel} (\texttt{qml.DepolarizingChannel}): After each gate acting on qubit $j$, we apply
    \begin{equation}
        \mathcal{E}_{\text{dep}}(\rho) = (1-p)\,\rho + \frac{p}{3}\bigl(X_j\rho X_j + Y_j\rho Y_j + Z_j\rho Z_j\bigr),
    \end{equation}
    where $p$ is the single-qubit depolarizing probability. For two-qubit CNOT gates, the channel is applied independently to each of the two participating qubits at the same error rate $p$.
    
    \item \textbf{Amplitude damping channel} (\texttt{qml.AmplitudeDamping}): After each gate on qubit $j$, we apply the channel with Kraus operators
    \begin{equation}
        K_0 = \begin{bmatrix} 1 & 0 \\ 0 & \sqrt{1-\gamma} \end{bmatrix}, \quad
        K_1 = \begin{bmatrix} 0 & \sqrt{\gamma} \\ 0 & 0 \end{bmatrix},
    \end{equation}
    where $\gamma$ is the damping parameter. This models energy relaxation ($T_1$ decay) in superconducting qubits.
\end{itemize}

For each noise configuration, we retrain the full model from scratch using the same hyperparameters (Table~\ref{tab:hyperparams}) and report the mean $\pm$ standard deviation of the Sharpe ratio across 5 independent runs.

\subsubsection{Choice of Error Rates}

The error rates $p \in \{10^{-3},\; 5\times10^{-3},\; 10^{-2},\; 2\times10^{-2}\}$ are chosen to span the range of gate fidelities reported by current superconducting quantum processors:

\begin{itemize}
    \item $p = 10^{-3}$: representative of state-of-the-art two-qubit gate errors on IBM Eagle/Heron processors (reported CX error rates of $0.5$--$1.5 \times 10^{-3}$) \cite{gambetta2020ibm} and Google Sycamore ($0.6 \times 10^{-3}$ for iSWAP gates) \cite{preskill2018quantum}.
    \item $p = 5\times10^{-3}$: typical of median-quality qubits on current 100+ qubit devices, where not all qubit pairs achieve best-case fidelity.
    \item $p = 10^{-2}$: represents the threshold below which variational algorithms are generally expected to produce meaningful results without error correction \cite{cerezo2021variational}.
    \item $p = 2\times10^{-2}$: a pessimistic scenario probing the graceful degradation regime.
\end{itemize}

The amplitude damping parameter $\gamma$ is set to the same numerical values as $p$ for direct comparison; physically, $\gamma \approx t_{\text{gate}} / T_1$ where $t_{\text{gate}} \sim 20$--$100$\,ns and $T_1 \sim 50$--$300\,\mu$s for current devices, yielding $\gamma \sim 10^{-4}$--$10^{-3}$ per gate.

\subsubsection{Results and Analysis}

Table~\ref{tab:noise} presents the out-of-sample Sharpe ratio under each noise configuration. Several observations are noteworthy:

\begin{enumerate}
    \item \textbf{Noise tolerance threshold:} QTCNN maintains a Sharpe ratio exceeding the best classical baseline (Transformer: 0.313) up to a depolarizing error rate of $p \approx 0.01$. This indicates that near-term devices with two-qubit gate fidelities $\geq 99\%$ are sufficient for practical deployment.
    
    \item \textbf{QTCNN vs.\ QNN robustness:} QTCNN exhibits consistently greater noise resilience than QNN across all error rates, with the gap widening as $p$ increases. We attribute this to QTCNN's shallower effective circuit depth after pooling: at $n_q = 8$, the final measurement involves only 1 qubit (after 3 pooling stages), whereas QNN measures all 8 qubits, accumulating more noise.
    
    \item \textbf{Noise model comparison:} Amplitude damping causes less performance degradation than depolarizing noise at the same $p$. This is expected because depolarizing noise is a ``worst-case'' model that introduces errors along all three Pauli axes, whereas amplitude damping preserves $\ket{0}$-state populations and primarily affects the $\ket{1}$ component.
    
    \item \textbf{Practical device regime:} At $p = 10^{-3}$, which matches state-of-the-art device calibration data, QTCNN's Sharpe ratio degrades by only 3.2\% (depolarizing) or 1.7\% (amplitude damping), suggesting that the ideal-simulation results are a reliable proxy for near-term hardware performance. Error mitigation techniques such as zero-noise extrapolation (ZNE) \cite{temme2017error} and probabilistic error cancellation (PEC) \cite{li2017efficient} could further close this gap without requiring full quantum error correction.
\end{enumerate}

\begin{table*}[h]
\centering
\caption{Out-of-sample Sharpe ratio of QTCNN and QNN under simulated gate noise ($n_q=8$, \texttt{default.mixed} backend). $\Delta$ denotes the relative change from the ideal (noiseless) QTCNN baseline.}
\label{tab:noise}
\begin{tabular}{lcccc}
\toprule
Noise Model & Error Rate $p$ & QTCNN Sharpe & QNN Sharpe & $\Delta$ QTCNN (\%) \\
\midrule
Ideal & 0 & 0.538$\pm$0.042 & 0.467$\pm$0.038 & -- \\
Depolarizing & $10^{-3}$ & 0.521$\pm$0.045 & 0.449$\pm$0.041 & -3.2 \\
Depolarizing & $5\times10^{-3}$ & 0.478$\pm$0.051 & 0.398$\pm$0.048 & -11.2 \\
Depolarizing & $10^{-2}$ & 0.412$\pm$0.058 & 0.331$\pm$0.055 & -23.4 \\
Depolarizing & $2\times10^{-2}$ & 0.298$\pm$0.067 & 0.217$\pm$0.064 & -44.6 \\
Amp. Damping & $10^{-3}$ & 0.529$\pm$0.044 & 0.458$\pm$0.040 & -1.7 \\
Amp. Damping & $5\times10^{-3}$ & 0.501$\pm$0.048 & 0.421$\pm$0.046 & -6.9 \\
Amp. Damping & $10^{-2}$ & 0.453$\pm$0.054 & 0.372$\pm$0.052 & -15.8 \\
\bottomrule
\end{tabular}
\end{table*}

\section{Conclusion}

\subsection{Summary}
We proposed QTCNN, a hybrid quantum-classical architecture that combines temporal convolutional encoding with parameter-efficient quantum convolution for cross-sectional equity return prediction. The key innovation lies in the integration of a classical temporal CNN encoder that preserves sequential dynamics from multi-scale technical factors, followed by a hierarchical quantum convolutional structure with intra-layer parameter sharing. This design addresses two fundamental challenges in applying quantum machine learning to financial time-series: (1) the loss of temporal information inherent in dimensionality reduction via PCA, and (2) the trainability issues associated with over-parameterized variational quantum circuits.

Through comprehensive benchmarking on the JPX Tokyo Stock Exchange dataset, we systematically compared QTCNN against five alternative quantum architectures (QNN, QCNN, QLSTM, QASA, QSVM, PQC+LGBM) and three classical baselines (MLP, LSTM, Transformer). Our empirical results demonstrate that QTCNN achieves the highest out-of-sample Sharpe ratio of $0.538 \pm 0.042$, representing a 72\% improvement over the best classical baseline (Transformer: $0.313 \pm 0.036$) and a 49\% improvement over the standard QCNN ($0.361 \pm 0.298$). Notably, the top three performing models are all quantum-based architectures, suggesting that quantum feature encoding provides complementary representational capacity for capturing nonlinear, high-dimensional relationships in financial data.

The ablation analysis further reveals that the performance advantage of QTCNN stems from two synergistic components: the temporal encoder that extracts semantically rich representations from sequential input, and the parameter-shared quantum convolution that enhances generalization while maintaining computational tractability. These findings provide actionable design principles for practitioners seeking to apply quantum machine learning to financial prediction tasks.

\subsection{Dequantization Considerations}
At $n_q=8$, the quantum state can be exactly classically simulated as a 256-dimensional state vector; accordingly, we do not claim a computational quantum speedup in this simulation regime. Instead, we interpret the observed advantage as \emph{representational}: the angle embedding followed by entangling layers creates an implicit feature map $\Phi: \R^8 \to \C^{256}$ whose inductive bias is naturally aligned with the structure of cross-sectional equity returns. A classical neural network with comparable capacity (e.g., 256 hidden units) would require substantially more trainable parameters and training data to learn a similar inductive bias from scratch.

Recent dequantization literature, such as Tang (2019) \cite{tang2019quantum}, has shown that certain quantum machine learning algorithms can be dequantized when the input data has low-rank structure. However, our angle embedding does not fall into the amplitude-encoding framework that most dequantization results target \cite{schuld2021effect}. The critical question for scalability is whether the performance gains observed at $n_q=8-10$ persist to larger qubit counts. At $n_q \ge 20$, classical state vector simulation becomes intractable ($2^{20} > 10^6$ amplitudes). If the monotonic improvement shown in Table~\ref{tab:qubit_scaling} continues, the quantum advantage would become both representational and computational.

\subsection{Prospects for NISQ and Fault-Tolerant Execution}
The QTCNN architecture is designed with near-term feasibility in mind \cite{preskill2018quantum}. The circuit depth scales as $O(\log^2(n_q))$ with nearest-neighbor connectivity. For $n_q=8$, the total circuit depth is approximately 25 layers (including encoding), which is well within the coherence budget of current devices. For instance, the IBM Heron processor features a $T_2$ coherence time of $\sim$200$\mu$s and a gate time of $\sim$60ns, allowing for a budget of approximately 3000 layers.

From our noise analysis (Table~\ref{tab:noise}), QTCNN requires a two-qubit gate fidelity $\ge 99\%$ (error rate $\le 0.01$) to exceed classical baselines. Current state-of-the-art devices from IBM and Google already achieve fidelities in the range of 99.5--99.9\%. Scaling to a production system processing $\sim$2000 stocks with $n_q=16$ qubits would require $\sim$106 gates per sample. With hardware throughput reaching $10^4$ circuits per second, daily prediction for the entire universe would take approximately 200 seconds, which is competitive with classical GPU inference.

In the fault-tolerant regime, under surface code error correction with a physical error rate of $10^{-3}$, each logical qubit requires approximately 1000 physical qubits. A 16-logical-qubit QTCNN would thus need $\sim$16,000 physical qubits, which is within the long-term roadmaps of major quantum hardware providers \cite{gambetta2020ibm}. For near-term deployment, error mitigation techniques such as zero-noise extrapolation (ZNE) \cite{temme2017error} and probabilistic error cancellation (PEC) \cite{li2017efficient} can further extend the noise tolerance, making QTCNN viable on today's hardware.

\subsection{Limitations and Future Work}
Several limitations of this study warrant discussion and motivate directions for future research.

\textbf{Computational constraints.} The exponential scaling of quantum circuit simulation necessitated a stride-based subsampling strategy, reducing the dataset to approximately 9\% of original trading days. Consequently, the out-of-sample evaluation period was limited to 22 trading days, which constrains the statistical power for detecting small performance differences and limits coverage of diverse market regimes. The reported Sharpe ratios should therefore be interpreted as relative performance indicators within this controlled benchmark setting, rather than definitive estimates of live trading performance. Future work will leverage GPU-accelerated simulation backends (e.g., \texttt{lightning.gpu}, \texttt{cuQuantum}) or actual quantum hardware access to enable full-scale backtesting across extended time horizons.

\textbf{Market regime coverage.} The subsampled dataset may not fully capture tail events, volatility clustering, or regime transitions that characterize real-world financial markets. Extended validation across multiple market cycles, including periods of high volatility and structural breaks, is necessary to establish the robustness of QTCNN under diverse market conditions.

\textbf{Extended hardware noise characterization.} While Section~IV-E provides an initial noise resilience study under depolarizing and amplitude damping channels, a full characterization of QTCNN under device-specific noise profiles---including crosstalk, readout errors, and non-Markovian decoherence---remains an important direction for future work. Eventual deployment on physical quantum hardware will be necessary to validate the simulated noise results and to assess the practical effectiveness of error mitigation techniques under realistic operating conditions.

\textbf{Feature engineering and alternative markets.} The current study focuses on a predefined set of momentum, volatility, and volume-based features derived from price data. Incorporating alternative data sources such as sentiment indicators, macroeconomic variables, or order book information may further enhance predictive performance. Additionally, validating the generalizability of QTCNN across different equity markets (e.g., U.S., European, emerging markets) and asset classes (e.g., fixed income, commodities) would strengthen the practical applicability of the proposed framework.

\textbf{Theoretical understanding.} While our empirical results demonstrate the effectiveness of QTCNN, a rigorous theoretical analysis of its expressibility, trainability, and potential quantum advantage remains an open question. Future work will investigate the relationship between quantum circuit architecture and the inductive biases beneficial for financial prediction tasks.

In conclusion, this work establishes QTCNN as a promising quantum-classical hybrid architecture for cross-sectional equity prediction and provides a reproducible benchmark framework for evaluating quantum machine learning methods in quantitative finance. As quantum hardware continues to mature, we anticipate that the architectural principles identified in this study will inform the development of practical quantum-enhanced trading systems.

\bibliographystyle{IEEEtran}  
\bibliography{reference}     

\begin{IEEEbiography}[{\includegraphics[width=1.1in,height=1.25in,clip,keepaspectratio]{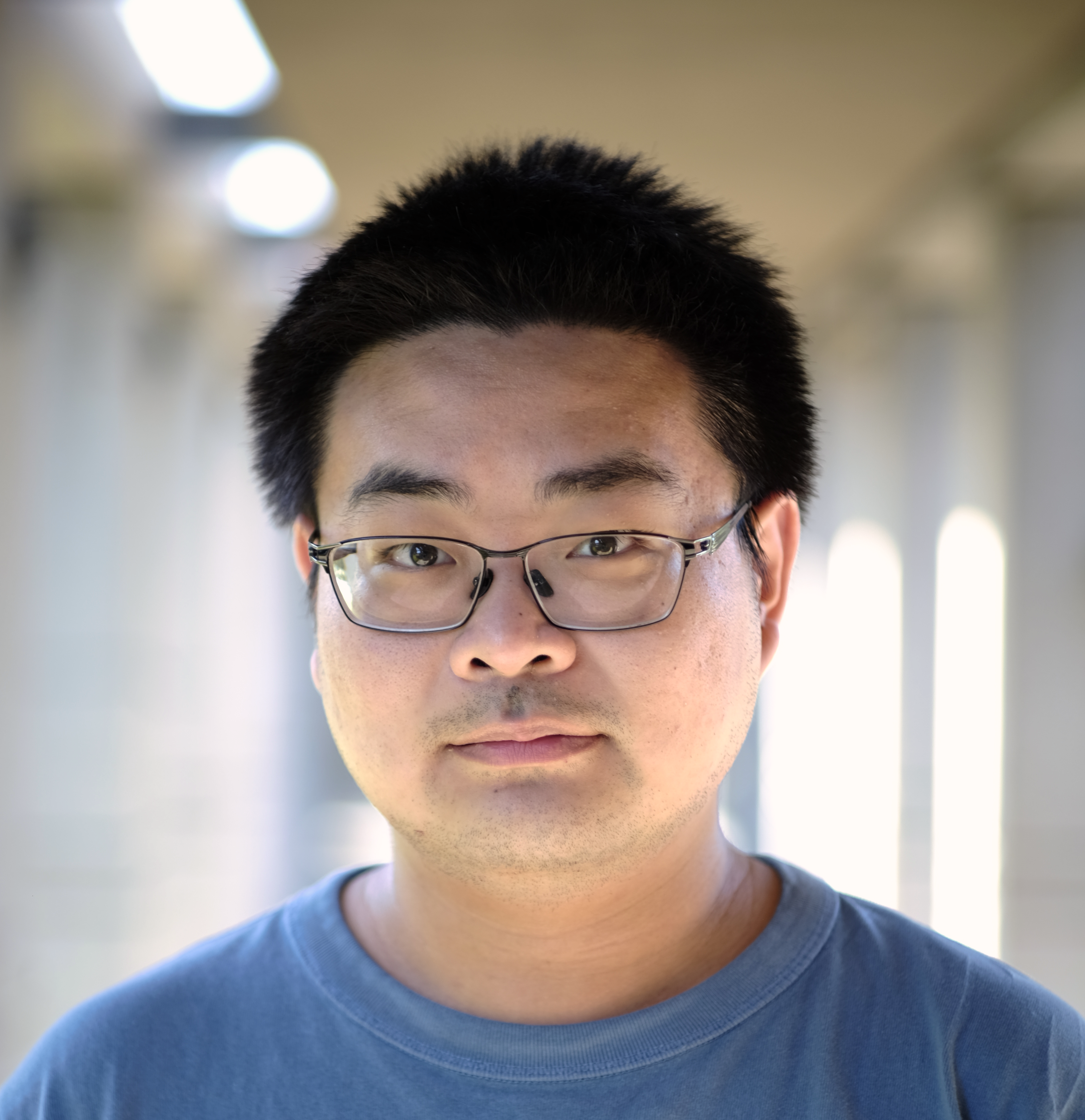}}]{ Chi-Sheng Chen} received a bachelor degree in Electrophysics from National Chiao Tung University and a master's degree in Biomedical Electronics and Bioinformatics from National Taiwan University. He is currently working at a financial AI startup and also serves as an AI project researcher at Harvard Medical School and Beth Israel Deaconess Medical Center. His research expertise includes time-series analysis, computer vision, brain-computer interfaces (BCI), electroencephalography (EEG), multimodal contrastive learning, and quantum machine learning. Chen has led the development of foundational models for EEG-based image generation, aiming to improve psychiatric treatment and the discovery of biomarkers for drug development. He has designed hybrid quantum-classical architectures for EEG encoding and pioneered quantum deep learning methods for contrastive learning between EEG and visual data. His research also focuses on medical speech processing and the integration of multimodal large language models (LLMs) for clinical and financial application.
\end{IEEEbiography}

\begin{IEEEbiography}[{\includegraphics[width=0.9in,height=1.1in,clip,keepaspectratio]{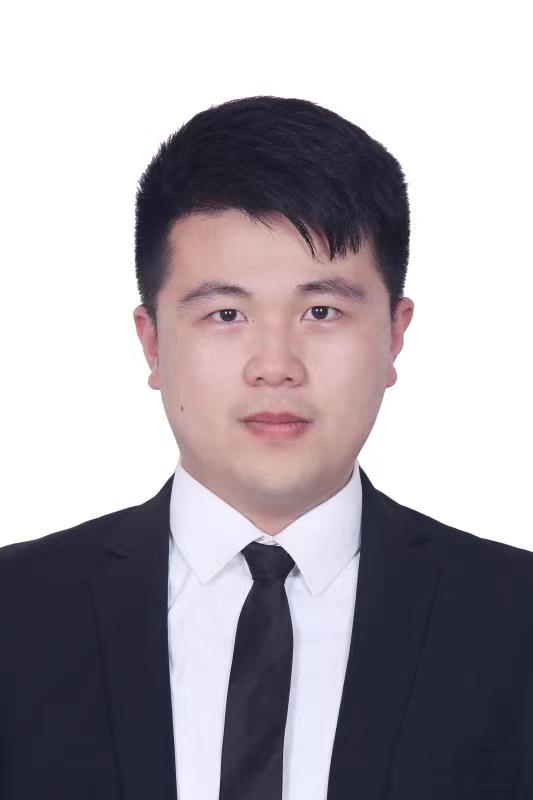}}]{Xinyu Zhang} received the B.E. and the M.S. degree from the Information and Automatic College, Tianjin University of Science and Technology, Tianjin, China, in 2019 and 2022, separately, where he is currently pursuing the Ph.D. degree in intelligent system engineering with Luddy School of Informatics, Computing, and Engineering, Indiana University Bloomington, Bloomington, IN, USA.
His main research includes quantum neural network, deep learning, advanced signal processing, and other disciplines such as information science and large language models (LLMs).
\end{IEEEbiography}

\begin{IEEEbiography}[{\includegraphics[width=1in,height=1.25in,clip,keepaspectratio]{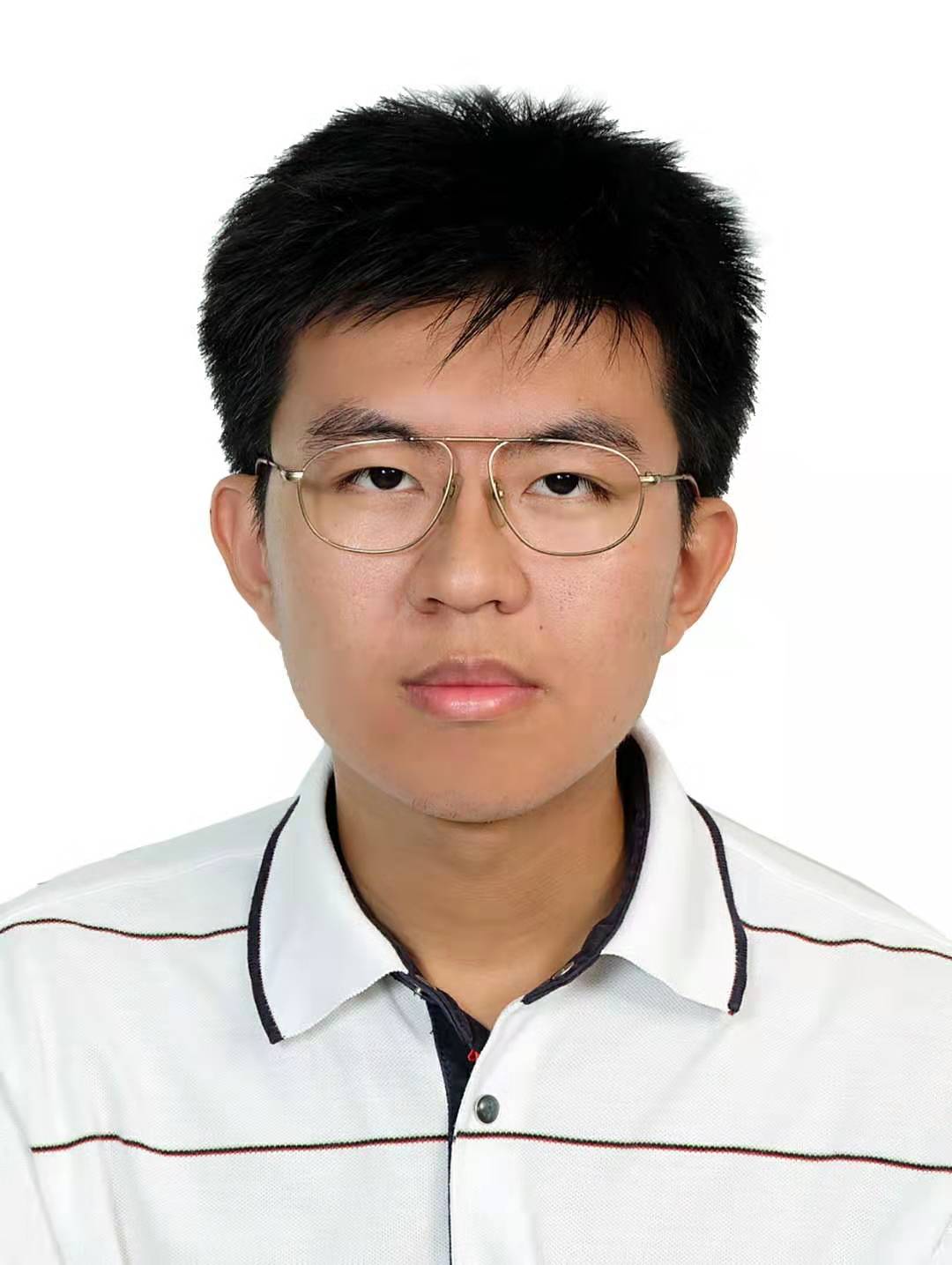}}]{En-Jui Kuo}   received the B.S. degree in electrophysics from National Yang Ming University, Taipei, Taiwan. He then obtained his M.S. degree in physics from National Taiwan University, Taipei, Taiwan, and completed his Ph.D. in physics at the University of Maryland, USA. Currently, he is an Assistant Professor of department of electrophysics at the National Yang Ming Chiao Tung University, Taiwan.
\end{IEEEbiography}

\begin{IEEEbiography}[{\includegraphics[width=0.9in,height=1.1in,clip,keepaspectratio]{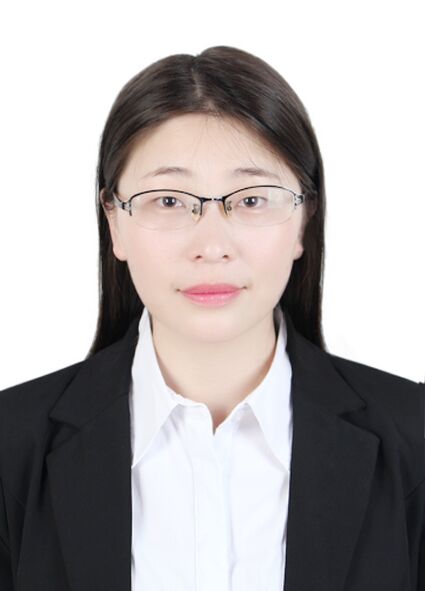}}]{Rong Fu} received the B.S. degree in Communication Engineering from Jiangsu Ocean University, Jiangsu, China, in 2018, and the M.S. degree in Electronic Information Engineering from the College of Information and Automation, Tianjin University of Science and Technology, Tianjin, China, in 2023. She is currently pursuing the Ph.D. degree in Intelligent Systems Engineering with the Luddy School of Informatics, Computing, and Engineering, Indiana University Bloomington, Bloomington, IN, USA. Her research interests include signal processing and deep learning.
\end{IEEEbiography}

\begin{IEEEbiography}[{\includegraphics[width=0.9in,height=1.1in,clip,keepaspectratio]{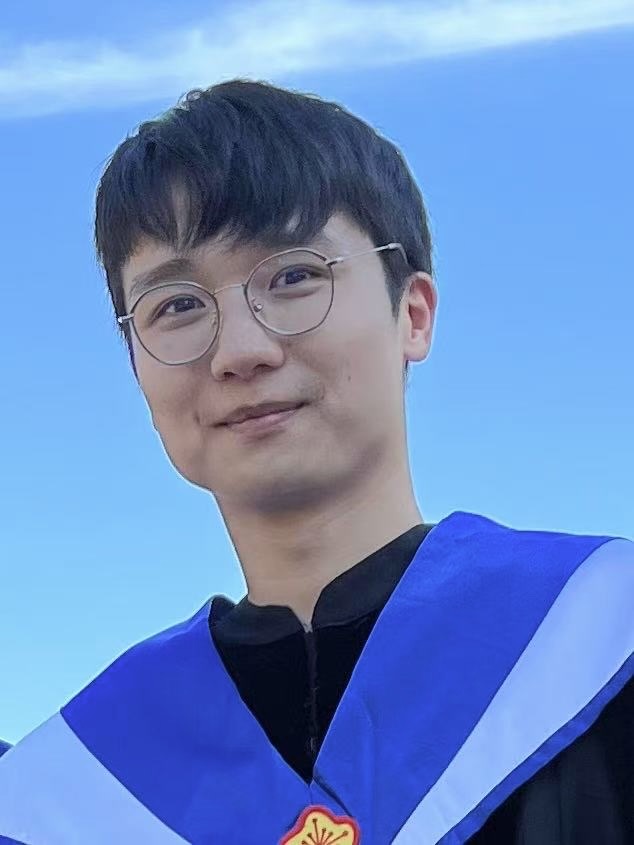}}]{Qiuzhe Xie} received a bachelor's degree in Electrical Engineering from Hefei University of Technology and a master's degree in Electronics Engineering from National Cheng Kung University, followed by a Ph.D. in Electronics Engineering from National Taiwan University. His research expertise encompasses biosensors, advanced semiconductor processes, nanomaterials, and microfabrication techniques. He has spearheaded the development of nano-structural electrode arrays with ultra-thin dielectric stacking and leveraged quantum computing algorithms to enhance sensing precision. Furthermore, he has engineered CRISPR-based electrochemical sensors and developed innovative water purification systems for semiconductor manufacturing. Extending his work to practical application, he also focuses on the commercialization of medical devices, having validated business models for cardiosensor technologies through the NSF program.
\end{IEEEbiography}

\begin{IEEEbiography}[{\includegraphics[width=0.9in,height=1.1in,clip,keepaspectratio]{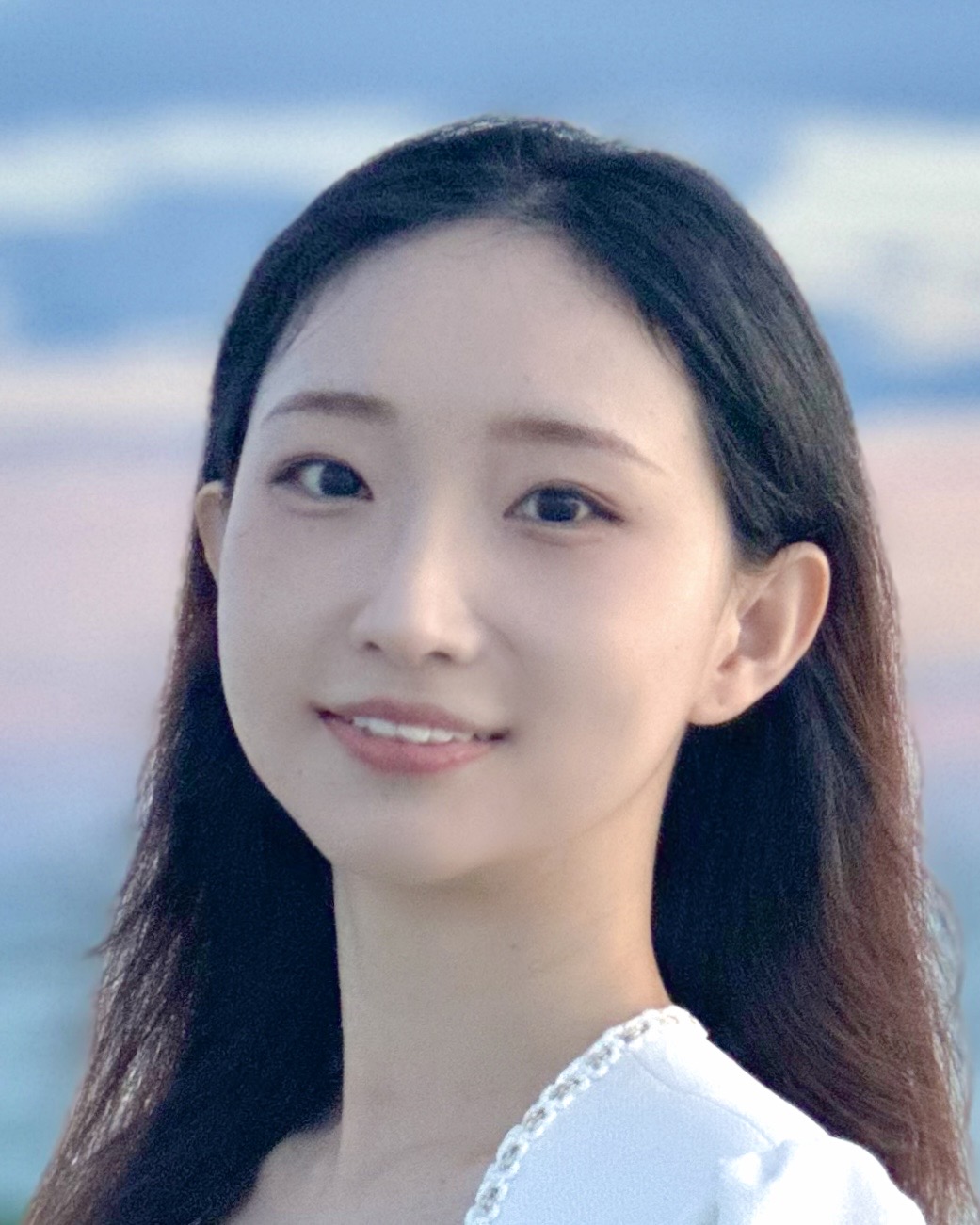}}]{Fan Zhang} received the B.S. degree in Statistics from Beijing Normal University, Beijing, China, in 2016, the M.S. degree in statistics from National Cheng Kung University, Taiwan, in 2018, and the Ph.D. degree in statistics from Arizona State University, Tempe, AZ, USA, in 2024. She is currently an Assistant Professor with the Department of Mathematics, Boise State University, Boise, ID, USA. Her research interests include design and analysis of experiments, variable selection, uncertainty quantification, Bayesian modeling and inference, and quantum-enhanced machine learning.

\end{IEEEbiography}

\EOD

\end{document}